\documentclass{article}

\usepackage{microtype}
\usepackage{graphicx}
\usepackage{booktabs} % for professional tables

\usepackage{hyperref}

\usepackage[accepted]{icml2024}

\usepackage{hyperref}
\usepackage{url}
\usepackage{mathtools}
\usepackage{caption}
\usepackage{hyperref}
\usepackage{algorithm}
\usepackage{algorithmic}
\usepackage{booktabs}
\usepackage{multirow}
\usepackage{soul}
\usepackage{xcolor}
\usepackage{xspace}
\usepackage{array}
\usepackage{lipsum}
\usepackage{subfigure}
\usepackage{amsmath, amsthm, amssymb}
\newcommand{\method}{\textit{FedAnchor}\xspace}
\usepackage{soul,color}
\usepackage{bbding}
\usepackage{makecell}

\usepackage[capitalize,noabbrev]{cleveref}

\theoremstyle{plain}

\theoremstyle{definition}

\theoremstyle{remark}

\usepackage[textsize=tiny]{todonotes}

\icmltitlerunning{FedAnchor: Enhancing Federated Semi-Supervised Learning with Label Contrastive Loss}

\begin{document}

\twocolumn[
\icmltitle{FedAnchor: Enhancing Federated Semi-Supervised Learning with Label Contrastive Loss for Unlabeled Clients}



\icmlsetsymbol{equal}{*}

\begin{icmlauthorlist}
\icmlauthor{Xinchi Qiu}{cam}
\icmlauthor{Yan Gao}{cam}
\icmlauthor{Lorenzo Sani}{cam}
\icmlauthor{Heng Pan}{cam}
\icmlauthor{Wanru Zhao}{cam}
\icmlauthor{Pedro P. B. Gusmao}{sur}
\icmlauthor{Mina Alibeigi}{zen}
\icmlauthor{Alex Iacob}{cam}
\icmlauthor{Nicholas D.\ Lane}{cam}

\end{icmlauthorlist}

\icmlaffiliation{cam}{Department of Computer Science and Technology, University of Cambridge}
\icmlaffiliation{sur}{Department of Computer Science, University of Surrey}
\icmlaffiliation{zen}{Zenseact}

\icmlcorrespondingauthor{Xinchi Qiu}{xq227@cam.ac.uk}

\icmlkeywords{Machine Learning, ICML, Semi-supervised Learning, Federated Learning}

\vskip 0.3in
]



\printAffiliationsAndNotice{}  % leave blank if no need to mention equal contribution

\begin{abstract}
Federated learning (FL) is a distributed learning paradigm that facilitates collaborative training of a shared global model across devices while keeping data localized. 
The deployment of FL in numerous real-world applications faces delays, primarily due to the prevalent reliance on supervised tasks. Generating detailed labels at edge devices, if feasible, is demanding, given resource constraints and the imperative for continuous data updates. In addressing these challenges, solutions such as federated semi-supervised learning (FSSL), which relies on unlabeled clients' data and a limited amount of labeled data on the server, become pivotal. In this paper, we propose \method, an innovative FSSL method that introduces a unique double-head structure, called \textit{anchor head}, paired with the classification head trained exclusively on labeled anchor data on the server. The \textit{anchor head} is empowered with a newly designed \textit{label contrastive loss} based on the cosine similarity metric. Our approach mitigates the confirmation bias and overfitting issues associated with pseudo-labeling techniques based on high-confidence model prediction samples. Extensive experiments on CIFAR10/100 and SVHN datasets demonstrate that our method outperforms the state-of-the-art method by a significant margin in terms of convergence rate and model accuracy. 

% \xq{emphazise the importance of using the anchor data, which makes the convergence faster}\ls{I've rephrase it a little bit trying to put more attention on the amazing results and the labeled anchor data}

\end{abstract}

\section{Introduction}\label{sec:intro}

% \xq{add a bit more motivations of the setups, and why low p-acc or low convergence rate can be harmful in the introduction.}

Federated learning (FL)~\citep{fedavg,flower} allows edge devices to collaboratively learn a shared global model while keeping their private data locally on the device, thereby decoupling the ability to do machine learning from the need to store data in the cloud. There are nearly seven billion connected Internet of Things (IoT) devices and three billion smartphones worldwide~\citep{lim2020federated}, potentially giving access to an astonishing amount of training data and decentralized computing power for meaningful research and applications. Most existing FL works focus on supervised learning, where the local private data is fully labeled. However, assuming that the full set of private data samples includes rich annotations is unrealistic for real-world applications~\citep{fedmatch, semifl,jin2020towards,yang2021federated,mao2023decentralized,viala2023towards}. 

% Although for some applications of FL, such as keyboard predictions, labeling data requires virtually no additional effort, in most cases, it requires the adoption of very difficult procedures that do not scale for massive populations of edge devices. 

\begin{figure}[t]
\centering
    \includegraphics[width=\linewidth]{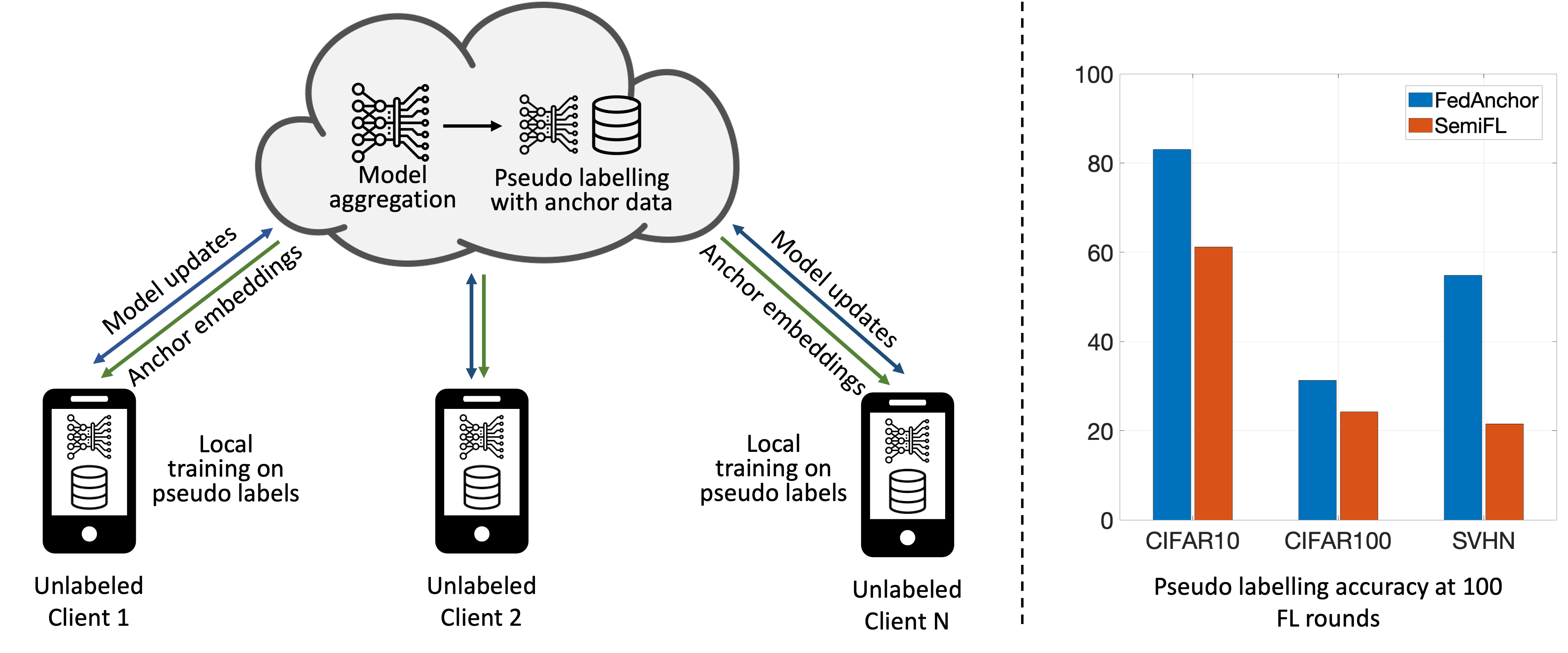}
\caption{\small (left) Pipeline of \method with pseudo labeling and anchor data on the server. 
% Anchor embeddings are only transmitted to the clients during downstream communication. 
(right) Pseudo labeling accuracy with 5000/2500/1000 anchor data on CIFAR10/CIFAR100/SVHN datasets, respectively.}
\label{fig:pip}
\vspace{-5mm}
\end{figure}

Acquiring large-scale labeled datasets on the user side can be extremely costly, e.g., many unlabeled data samples are generated through interactions with smart devices in daily life, such as pictures or physiological indicators. % measured by wearable devices.
These volumes of data make it impractical to mandate individual users to annotate the data manually.
This task can be excessively time-intensive for users, or they may lack the requisite advanced knowledge or expertise for accurate annotation, particularly when the dataset pertains to a specialized domain such as medical data~\citep{yang2021federated} or autonomous driving \cite{elbir2020federated}.
The complicated process of annotation results in most user data remaining unlabeled, further preventing the conventional FL pipeline from conducting supervised learning.
Recent studies of self-supervised learning in FL attempt to leverage unlabeled user data to learn robust representations~\citep {gao2022federated,rehman2022federated,rehman2023dawa}.
However, the learned model still requires fine-tuning on labeled data for downstream supervised tasks.

Compared to FL environments, centralized data annotation is more straightforward and precise in data centers. Even in low-resource contexts (e.g., medical data), the task of labeling limited data stored on the server by experts would not demand substantial effort. The integration of semi-supervised learning (SSL)~\citep{chapelle2009semi, yang2022survey,fixmatch,mixmatch} with FL can potentially leverage limited centralized labeled data to generate pseudo labels for supervised training on unlabeled clients. Existing work~\citep{jeong2020federated,zhang2021improving} attempted to perform SSL by using off-the-shelf methods only, such as FixMatch~\citep{fixmatch}, MixMatch~\citep{mixmatch} in FL environments. Although these methods provide certain model convergence guarantees during the FL stage, they cause heavy traffic for data communication due to their per-mini-batch communication protocol. Another recent work, SemiFL~\citep{semifl}, improves the training procedure by implicitly conducting entropy minimization. This is achieved by constructing hard (one-hot) labels from high-confidence predictions on unlabeled data using existing centralized SSL methods. These are subsequently used as training targets in a standard cross-entropy loss. However, it is argued that using pseudo-labels based on model predictions might lead to a confirmation bias problem or overfitting to easy-to-learn data samples~\citep{nguyen2023boosting}. In addition, the existing methods typically establish a pre-defined threshold, generally set relatively high to only keep the high-confident pseudo-labels. This might lead to slow convergence issues, especially at the beginning of training when very limited samples satisfy the threshold.

In this paper, we propose an enhanced federated SSL method, dubbed \method ~- a newly designed label contrastive loss based on the cosine similarity metric to train on labeled anchor data on the server. Instead of retaining the high-confidence data solely through model prediction in the conventional SSL studies, \method \textit{for-the-first-time} generates the pseudo labels by comparing the similarities between the model representations of unlabeled data and labeled anchor data. This provides better-quality pseudo-labels, as shown in Fig.~\ref{fig:pip} (right), which alleviates the confirmation bias and reduces the issues of over-fitting to easy-to-learn data samples. Our contributions are summarized as follows: 1) we propose a \textit{unique} pseudo-labeling method \method for SSL in FL, which leverages the similarities between feature embeddings of unlabeled data and labeled anchor data; 2) we design a novel label contrastive loss to improve the quality of pseudo labels in FSSL further; 3) we perform extensive experiments on three representative datasets having different amounts of labeled anchor data and show that the proposed methods achieve the state-of-the-art (SOTA) performance with a faster convergence rate. %The exact implementation can be found in the anonymous link \footnote{\url{https://anonymous.4open.science/r/fedanchor-8727/README.md}}.
%\ls{cite flower}

\section{Background}\label{sec:related}
\vspace{-1mm}

\paragraph{Federated Learning (FL).} FL aims to collaboratively learn a global model while keeping private data on the devices. We consider a $C$ class classification problem defined over a compact space $\mathcal{X}$ and a label space $\mathcal{Y} = [C]$, where $[C] = \{ 1,...,C \}$. The FL training objective is to minimize: $\min_w f(w) = \sum_{k=1}^N p_k F_k(w) = \mathbb{E}_k \big(F_k(w)\big)$, where $N$ is the total number of clients in the client pool, $p_k > 0$ and $\sum p_k = 1$. In general, the local objectives measure the local empirical risk over possibly differing data distribution $\mathcal{D}_k$ of each client, i.e., $F_k(w):= \mathbb{E}_{x_k \sim \mathcal{D}_k} \big(f_k(w;x_k)\big)$, with $n_k$ samples available at each client k. The standard FedAvg \cite{fedavg} method set $p_k = \frac{n_k}{n}$ where $n =  \sum_k n_k$ is the total number of data points in the systems. $f$ is parameterized over the hypothesis class $w$, which can be seen as the weight of the neural network. $F_k$ is the loss function specific to the task. 

\paragraph{Semi-Supervised Learning (SSL).} SSL is a problem of learning with partially labeled data, especially when the amount of unlabeled data is much larger than labeled ones \citep{zhou2005tri, rasmus2015semi}. The standard SSL method involves giving pseudo-labels to unlabeled data \citep{lee2013pseudo} and using these pseudo-labels as hard labels for supervised training. On the other hand, consistency regularization~\citep{bachman2014learning} methods train models by minimizing the distance among stochastic outputs, which can be achieved through different weak or strong augmentations~\citep{cubuk2020randaugment, thulasidasan2019mixup, french2017self}. Methods such as MixMatch~\citep{mixmatch} and FixMatch~\citep{fixmatch} combine both ideas by imposing a threshold on the model predictions of weak and strong augmented samples, retaining artificial labels only for those with the most significant class probability falling above a pre-defined level.

\paragraph{Federated Semi-Supervised Learning (FSSL).} 
Considering the difficulties of labeling data in a federated setting, FSSL represents the federated variant of SSL. In this context, we address the more challenging yet realistic scenario where the data stored on the client side is completely unlabeled. Given a dataset $\mathcal{D} = \{ \textbf{x}_i,y_i \}^N_{i=1}$, which consists of both a labeled set $\mathcal{S}= \{ \textbf{x}_i,y_i \}^S_{i=1}$, named \textit{anchor data} in our paper, and an unlabeled set $\mathcal{U}= \{ \textbf{x}_i \}^U_{i=1}$. The unique challenge of FSSL comes from the fact that taking off-the-shelf SSL methods and applying them to FL cannot achieve communication-efficient FL training. For example, techniques such as FixMatch~\citep{fixmatch} or MixMatch~\citep{mixmatch} require each mini-batch to sample from both labeled and unlabeled data samples with a carefully tuned ratio, which is impossible to achieve in real FL settings if the labeled data and unlabeled data are stored in different places. FedMatch~\citep{fedmatch} splits model parameters for labeled servers and unlabeled clients separately. FedRGD~\citep{zhang2021improving} trains and aggregates the model of the labeled server and unlabeled clients in parallel with the group-side re-weighting scheme while replacing the batch normalization to group normalization.
%\cite{liu2021federated} and \cite{saha2023rethinking} considered the case when there are both labeled and unlabeled data at clients in the real medical settings; imFed-Semi \cite{jiang2022dynamic} conducts the client training by exploiting class proportion information; FedoSSL \cite{zhang2023towards} proposes a framework tackling the biased training process for heterogeneously distributed and unseen classes; FedSiam~\cite{long2020fedsiam} integrates a siamese network into FL, incorporating a momentum update to address the non-IID challenges posed by unlabeled data; FedIL~\cite{yang2023fedil} employs iterative similarity fusion to maintain consistency between server and client predictions on unlabeled data, and utilizes incremental confidence to construct reliable pseudo-labels.

FedCon \cite{long2021fedcon} borrows from BYOL \citep{grill2020bootstrap} the idea of using two models and proposes using a \emph{consistency loss} between two different augmentations to help clients' networks learn the embedding projection. 
SemiFL~\citep{semifl} takes the centralized SSL method MixMatch~\citep{mixmatch} with mixup \citep{thulasidasan2019mixup} augmentation method together with an alternate training scheme~\citep{gao2022end,dimitriadis2020federated} to achieve the current SOTA performance. 
However, the simple combination of existing centralized SSL methods with different augmentation approaches in SemiFL insufficiently improves the quality of pseudo-labeling.
% \xq{It is worth noting that the pseudo-labeling and the mixup method implemented in the SemiFL are adopted by other methods, too, and the different augmentation methods can also work together with the SemiFL training pipeline.} 
Both FedCon and SemiFL serve as baselines in our paper.
% \wz{We can emphasize the shortcomings of these previous methods compared to our method here.}

\vspace{-2mm}
\paragraph{Latent Representation.} The great success neural networks have achieved since their introduction is to be adjudged to their capability of learning latent representations of the input data that can eventually be used to learn the task they have been designed for.
The set containing this latent information is often referred to as \textit{latent space}, which can also be the subject of topological investigation~\cite{zaheer2017deep, hensel2021survey}.
Many theoretical studies have highlighted the importance of such representations as they are explicitly identified in many settings, e.g., the intermediate layers of a ResNet architecture~\cite{he2016deep}, the word embedding space of a language model, or the bottleneck of an Autoencoder~\cite{moor2020topological}. More recently, the research community focused on investigating the \textit{quality} of the \textit{latent space} as well-performing networks have shown similar learned representations~\cite{li2015convergent, morcos2018insights, kornblith2019similarity, tsitsulin2019shape, vulic2020all}.
Despite these speculations being found to be more empirical than theoretical, the interest in leveraging latent representation to enhance or facilitate training methods, especially those not relying on labeled data, has increased.

The most audacious attempt to leverage the structure of different latent spaces is presented in~\cite{morcos2018insights}.
The authors of the latter leveraged interesting observations regarding the structure of different latent spaces learned by diverse training procedures to achieve \textit{zero-shot model stitching}. While we refer the reader to their paper for the details, we want to highlight one observation and one method that makes their work relevant to ours.
First, they showed that differently-learned latent spaces are the same up to an approximately isometric transformation.
Second, they used the latter observation to construct a method based on ``anchor" samples and the ``cosine similarity" metric.
% Such analysis was never extended to a semi-supervised learning or federated learning context.

\section{Motivation}\label{sec:keyobservations}

Most current SSL methods, as mentioned in Section \ref{sec:related}, are based on augmentation methods and pseudo-labels obtained from the prediction of the training model. This method not only gives confirmation bias and overfitting to easy-to-learn data samples~\citep{nguyen2023boosting}, but it also makes the number of training samples qualified for training very small, i.e., an unlabeled sample will only be considered for pseudo-labeling when the probability of a given class is above a pre-defined and difficult-to-tune confidence threshold (often set as $0.95$ for high confidence).
% Since the quality of the pseudo-labels is crucial for a good final performance, previous methods rely on a pre-defined and difficult-to-tune confidence threshold to select the samples to address with pseudo-labels.
% An unlabeled sample will only be considered for pseudo-labeling when the probability of a given class is above the threshold (often set as $0.95$ for high confidence).
% The prediction of the training model is used as the pseudo-labels for further training.
% However, this method not only gives confirmation bias and overfitting to easy-to-learn data samples~\citep{nguyen2023boosting}, but it also makes the number of training samples qualified for training very small.
% Thus, we designed \method to be robust against uncareful tuning of any confidence threshold.

We now present preliminary investigations on the current SOTA method, SemiFL, to showcase these shortcomings. We run these investigations on the CIFAR10 dataset \cite{krizhevsky2009learning} under the FL settings when it is partitioned non-IID ($\alpha = 0.1$) over $100$ clients, with $10$ clients selected for training during every communication round. More details of the experimental protocols can be found in Section \ref{sec:exp_setup}. In this section, we demonstrate two different setups with different anchor sizes, $250$ and $500$.

\begin{figure}[t]
\centering
    \includegraphics[width=\linewidth]{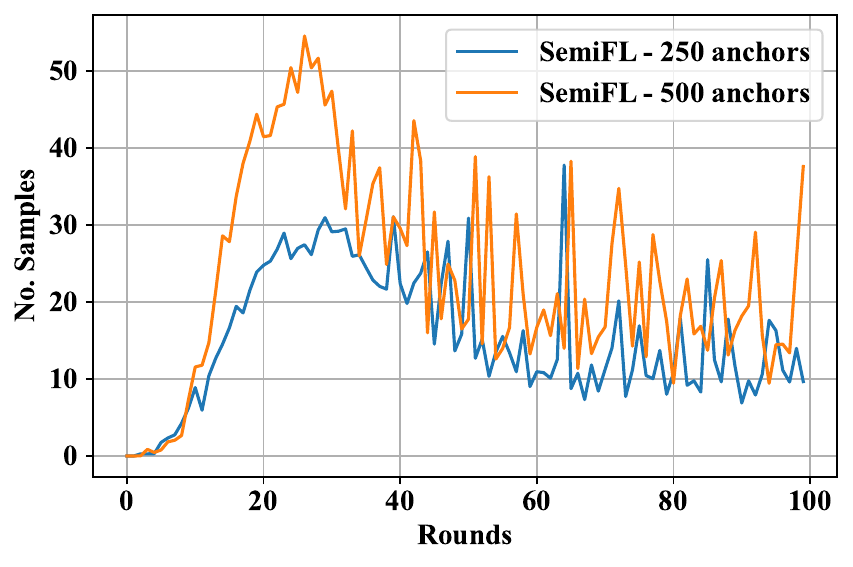}
\caption{\small The average number of qualified data samples trained by selected clients in each round, for CIFAR10 Non-IID ($\alpha=0.1$) partition with $250$/$500$ anchor data on the server, using method SemiFL. The plot shows that the average number of data samples trained by clients is very small, given each client has a total of $500$ training data samples.}
\label{fig:semi_no_samples}
\vspace{-4mm}
\end{figure}

Figure \ref{fig:semi_no_samples} demonstrates that the average number of samples per client satisfying the threshold condition is very low.
For the CIFAR10 dataset, every client has $500$ data samples after the partition, but as we can see from the figure, the maximum number of samples satisfying the condition is below $60$, and on average, it is below $25$ for anchor size of $500$ and below $20$ for the anchor size of $250$.
Therefore, the training procedure cannot efficiently take advantage of the information available, resulting in a slower convergence rate.
This training method incurs unnecessary communication costs, which can be more energy-consuming, as shown in previous work \citep{qiu2023first}.

Our other baseline, FedCon~\citep{long2021fedcon} shows another shortcoming: choosing the best augmentation for the task is crucial for its consistency loss to work. Moreover, in addition to tuning standard FL-related hyperparameters, FedCon requires careful tuning of several local and global optimization steps. The flexibility and robustness of \method overcome the shortcomings of previous proposals that rely too much on difficult-to-tune hyperparameters and specific augmentations.

The quality of the pseudo-labels generated by the \method allows for overcoming previous works' shortcomings while using the same amount of samples on the server. Thus, a potential application in the wild to a more extensive federated population will require the construction of a smaller dataset at the server to serve as \emph{anchor} compared to the baselines.
The model achieves a better convergence rate and performance using our more efficient training procedure using the same number of anchor samples. 

\section{Methodology: \method}\label{sec:method}
In this section, we present our method \method (Fig.~\ref{fig:pip}) in detail. \method aims to fully utilize the information embedded in the anchor dataset stored on the server to provide better pseudo-labels for unlabeled private client's data to be trained on supervised tasks. We propose a novel label contrastive loss combined with cosine similarity metrics to extract the anchor information in the latent space. The section is organized as below: we detail the new label contrastive loss in Section \ref{sec:newloss}; we then explain in Section \ref{sec:pseudolabel} how to obtain the pseudo label in our methods; then it is followed by the algorithm of local training on the client side in Section \ref{sec:localtraining} and the server training using anchor data in Section \ref{sec:servertraining}. The pseudo-code of the \method is summarized in Algorithm \ref{algorithm:algo}.

We define the $x^m$ to be the unlabeled data samples on client $m$; $(x^{anchor}, y^{anchor})$ be the labeled anchor data; $z^m$ be the output of the \textit{anchor head} (latent space) for the unlabeled client data, and $z^{anchor}$ be the output of the \textit{anchor head} (latent space) for the anchor data. Hence, the symbol $z$ always represents any given data sample's latent representation. Also, $\hat{y}$ denotes the pseudo-label for unlabeled data.

\subsection{Label contrastive loss}\label{sec:newloss}
One of the main novelties of our method lies in introducing a new label contrastive loss that acts on the latent space. As mentioned in the background section (Section \ref{sec:related}), the latent space retains important representations that can be identified and utilized in many settings. In our case, the latent space can be the output of any pre-defined layer in the neural network. However, since our method modifies the underlying geometry of the chosen output layer, we suggest picking a later layer to let the earlier ones maintain their feature extraction capabilities. A visualization can be found in Appendix \ref{app:latent_space}.

\method uses a double-head structure for the model, with one head (\textit{classification head}) consisting of the original classification layers and the other (\textit{anchor head}) consisting of a projection layer to the latent space to reduce the previous embedding dimension. The \textit{anchor head} is crucially designed to work with the new label contrastive loss that we propose below. Despite its complexity, this structure is model agnostic and can be implemented in all the deep learning model architectures used for a classification task. 

The label contrastive loss aims to map the $z^{anchor}$ of the same label to the same local region in the latent space while forcing data with different labels to be far from each other. We let $s_{i, j}$ be any similarity function between data sample $x_i$ and $x_j$ in the latent space ($z_i$,$z_j$). Cosine similarity is used in our case: $ s_{z_i, z_j} = 	\frac{z_i ^T z_j}{\lVert z_i \rVert \lVert z_j \rVert}$. 

Then, we propose our new label contrastive loss as in eq.~\ref{eq:cossimloss} and~\ref{eq:cossimloss2}. The label contrastive loss is defined on a batch of anchor data samples. Given a batch of anchor data, we calculate the $l(c)$ for each label class, then sum up all label classes to obtain the final value of the label contrastive loss. 

\vspace{-2mm}
\begin{align}
    l(c) = -\log &\frac{\sum _{y(i), y(j) = c}\exp (s_{z_i,z_j}/\tau)}{\sum_{y(i) \not = y(j) } \exp(s_{z_i,z_j}/\tau)}  \label{eq:cossimloss} \\
    \mathcal{L}_c &=  \frac{1}{C} \sum_{c=1}^C l(c) \label{eq:cossimloss2}
\end{align}    
where $\tau$ is the tunable temperature hyper-parameter. During the training, the label conservative loss tends to maximize the similarities between samples with the same labels while minimizing the similarities between samples with different labels. Crucially, the proposed loss will correct the projections of the data samples by modifying the underlying geometry of the projector part of the model.

\subsection{Pseudo-labeling using \textit{anchor head}} \label{sec:pseudolabel}
We present our pseudo-labeling method by describing a single communication round, the unit of iteration in FL, step-by-step. At the communication round $t$, the server selects a subset of clients to participate with participation ratio $r$ in the current federated round. The server will then broadcast the model parameters ($w_t$) and the latent anchor representations \big($\{ z^{anchor}_i \}_{i=1}^S$\big) to the selected clients. The anchor latent representations are the output of the \textit{anchor head}, which generates the hard pseudo-label during training. 

After receiving the current model weights and anchor latent representations, each client $m$ computes the latent representation of each local data \big($\{ z^m_i \}_i$\big). Then, it compares $z^m_i$ to each anchor latent representation \big($\{ z^{anchor}_i \}_{i=1}^S$\big) to obtain the pseudo-label leveraging the cosine similarity between $z^m_i$ and each anchor latent representation. The scores obtained from this comparison are averaged by label, i.e.,~cosine similarities relative to anchor latent representation from anchor samples with the same label are averaged. The pseudo-label is the label that provides the maximum score.

Let $Z^{anchor}_{c} = \{ z^{anchor}_{i} \in S~| \ y^{anchor}_i = c \}$ be the set of anchor latent representation with label $c$. Then, the average cosine similarities between $z^m_i$ and each $Z_c$ can be computed. Let $s^{avg}$ be the average similarities of a data input compared with anchor data. $s^{avg}_c(z^m_{i})$ is the average cosine similarities between unlabeled data latent representation $z^m_{i}$ compared with each anchor data with label class $c$. The value of $s^{avg}_c(z^m_{i})$ can be calculated as:

\begin{equation}\label{eq:savg}
   s^{avg}_c(z^m_{i}) = \frac{1}{|Z^{anchor}_{c}|} \sum_{z_k \in Z^{anchor}_{c}} s(z^m_{i}, z_k) ,
\end{equation}

Subsequently, to obtain the pseudo-label $\hat{y}^m_{i}$ for each local data, we need to find the label class that produces the maximum $s^{avg}_c(z^m_{i})$:
\vspace{-2mm}
\begin{equation}\label{eq:yhat}
    \hat{y}^m_{i} = argmax_{c \in [C]} \big( s^{avg}_c(z^m_{i}) \big).
\end{equation}
%where  $\hat{y}$ denotes the pseudo-label.

It is worth noting that cosine similarity is well-known for being low complexity and highly optimizable \cite{novotny2018implementation}. Such operations are independent and completely parallelizable. As a result, it is feasible to scale up to cases where clients and server contains a large amount of data.

\subsection{Local training} \label{sec:localtraining}
After obtaining the pseudo-labels ($\hat{y}^m$) for the local data, each client will locally perform supervised training using the pseudo-labels as hard labels.
As mentioned in Section \ref{sec:related}, semi-supervised training methods often leverage different data augmentation procedures -- weak or strong augmentation methods, such as RandAugment~\citep{cubuk2020randaugment}.
These perform differently depending on the dataset, model, and task.
We used the \textit{mixup}~\citep{thulasidasan2019mixup} method to obtain a fair comparison with SemiFL. The \textit{mixup} method trains a neural network on convex combinations of pairs of examples and their labels with coefficients generated by the beta distribution. This can potentially improve the robustness of the model and utilize the limited labeled data. It is important to notice that \method is completely agnostic on the augmentation used, as opposed to FedCon, whose consistency loss's performance strongly depends on the specific augmentation. The detailed implementation of mixup methods can be found in Appendix \ref{app:mixup}.

\vspace{-2mm}
\subsection{Server training} \label{sec:servertraining}
We aim to use the labeled anchor data on the server as training data for supervised and label contrastive loss to leverage the information they carry fully.
Training at the server on the labeled data is not novel \cite{gao2022end, dimitriadis2020federated} as SemiFL \citep{semifl} performs this dubbing \textit{alternate training} procedure. However, by comparison, \method trains on the anchor data at the server in two epochs: one for the supervised classification loss and one for the label contrastive loss to further improve the pseudo-label accuracies during every round. Let $\mathcal{L}_s$ be the supervised training loss, such as the standard cross-entropy loss for the classification task, and $\mathcal{L}_c$ be the label contrastive loss described in Section \ref{sec:newloss}. Therefore, the server will train for one epoch on the \textit{classification head} by minimizing the loss $\mathcal{L}_s$ and for one epoch on the \textit{anchor head} by minimizing the loss $\mathcal{L}_c$. %Therefore, server training on the anchor data will be the training on the supervised loss $\mathcal{L}_s$ followed by the training on the label contrastive loss $\mathcal{L}_c$.

\begin{table*}[ht!]
    \caption{\small Comparison of \method with the state-of-the-art methods and fully supervised baseline on model ResNet-18. \method \textit{(mix)} represents the experiments conducted using \method, but the supervised training on the server with anchor data is replaced with the \textit{mixup} method explained in Section~\ref{sec:extension}. 
    Supervised models are trained with standard FL procedure with alternative training on the server, initialized from a pre-trained model on anchor data. 
    % \yg{need to change after new supervised results, also make sure to say that supervised is also running alternative training , but server without mixup method}
    % ``Pre-trained" represents the models trained with anchor data in prior to FL. 
    }
    \vspace{1mm}
    \label{tab:main}
    \centering
    \renewcommand{\arraystretch}{1.5}
    \scalebox{0.675}{
    \begin{tabular}{lc ccc cc cc}
    \toprule
    \multicolumn{2}{l}{Datasets} & \multicolumn{3}{c}{CIFAR10} & \multicolumn{2}{c}{CIFAR100} & \multicolumn{2}{c}{SVHN} \\
    \hline
    \multicolumn{2}{l}{Number of anchor data} & 250 & 500 & 5000 & 2500 & 10000 & 250 & 1000 \\
    \hline
    % \multicolumn{2}{l}{Pre-trained} &  &  &  &  &  &  &  \\
    % \hline
    \multirow{6}{*}{IID ($Dir=1000$)} & Supervised & 89.45 ± 0.47 & 89.73 ± 0.09 & 89.07 ± 0.22 & 61.84 ± 0.17 & 63.33 ±  0.21& 95.38 ± 0.03 & 94.87 ± 0.53 \\
    \cline{2-9}
    & FedCon & 34.94 ± 0.43 & 50.81 ± 3.21 & 74.95 ± 1.26 & 32.84 ± 0.40 & 50.05 ± 0.34 & 54.83 ± 2.77 & 83.92 ± 1.03 \\
    & FedAvg+FixMatch & 33.98 ± 1.77 & 49.18 ± 2.33 & 75.42 ± 0.73 & 32.31 ± 0.83 & 49.15 ± 0.57 & 43.61 ± 0.64 & 81.65 ± 1.83 \\
    & SemiFL & 77.82 ± 0.49 & 81.19 ± 0.35 & 75.46 ± 0.19 & 48.20 ± 0.63 & 63.68 ± 0.16 & 91.55 ± 0.77 & 90.11 ± 1.17 \\
    & FedAnchor & 80.36 ± 0.18 & \textbf{85.94 ± 0.11} & 83.52 ± 0.41 & 50.79 ± 0.27 & 62.02 ± 0.24 & \textbf{91.74 ± 0.41} & \textbf{92.77 ± 0.11} \\
    & FedAnchor (\textit{mix}) & \textbf{82.82 ± 0.21} & 85.87 ± 0.25 & \textbf{84.43 ± 0.36} & \textbf{51.34 ± 0.07} & \textbf{63.99 ± 0.39} & 87.46 ± 0.63 & 92.71 ± 0.54 \\
    \hline
    \hline
    \multirow{6}{*}{Non-IID ($Dir=0.1$)} & Supervised & 75.42 ± 5.64 & 77.96 ± 2.55 & 77.99 ± 1.24 &  50.87 ± 1.64 & 60.47 ± 0.52& 87.48 ± 4.78 & 91.29 ± 0.33 \\
    \cline{2-9}
    & FedCon & 38.46 ± 0.42 & 51.57 ± 1.34 & 76.38 ± 1.36 & 32.00 ± 0.46 & 48.61 ± 0.56 & 50.86 ± 1.50 & 83.40 ± 1.89 \\
     & FedAvg+FixMatch & 39.10 ± 0.17 & 49.92 ± 2.49 & 73.17 ± 1.33 & 34.43 ± 0.87 & 49.53 ± 0.56 & 47.09 ± 1.31 & 76.83 ± 3.26 \\
    & SemiFL & 58.82 ± 0.72 & 68.96 ± 0.98 & 72.12 ± 0.35 & 42.41 ± 0.47 & 59.72 ± 0.31 & 68.97 ± 13.24 & 87.21 ± 1.66 \\
    & FedAnchor & 60.19 ± 0.32 & 72.75 ± 0.63 & 81.37 ± 0.31 & 43.50 ± 0.13 & 59.96 ± 0.40  & \textbf{77.42 ± 0.55} & \textbf{90.20 ± 0.56} \\
    & FedAnchor (\textit{mix}) & \textbf{62.94 ± 0.52} & \textbf{73.02 ± 0.31} & \textbf{83.59 ± 0.46} & \textbf{46.39 ± 0.36} & \textbf{61.01 ± 0.06} & 60.30 ± 5.34 & 87.28 ± 0.08 \\
    \bottomrule
   \end{tabular}  
   }
\end{table*}

\subsection{Possible Additions} \label{sec:extension}
Since the assumptions we made to design \method are simple and general, numerous potential extensions can be added to the above-described pipeline. One addition can be made to the supervised training on the server. Instead of only using strong augmentation, we can implement the same idea of mixup augmentation with a loss function. %To refer to our discussion above, the \textit{fix dataset} will be the full anchor dataset, and we sample from the full anchor dataset with replacement to form the \textit{mix dataset}. 

Additionally, at the pseudo-labeling stage (Section \ref{sec:pseudolabel}), instead of feeding the raw and original unlabeled training data, we can borrow the idea of consistency regularization and pseudo-label ensembles techniques \citep{bachman2014learning, sajjadi2016regularization} to weakly augment the unlabeled training data \big($\alpha(x^m_{i})$\big) for a few times and then take the ensembles to generate more robust pseudo-labels.

\section{Experiments}\label{sec:exp}

\begin{table*}[ht!]
    \caption{\small Comparison of \method with the state-of-the-art methods and fully supervised baseline on model Wide-Resnet28x2. 
    % \method \textit{(mix)} represents the experiments conducted using \method, but the supervised training on the server with anchor data is replaced with the \textit{mixup} method explained in Section~\ref{sec:extension}. 
    % Supervised models are trained with standard FL procedure with alternative training on the server, initialized from a pre-trained model on anchor data. 
    }
    % \method \textit{(mix)} represents the experiments conducted using \method, but the supervised training on the server with anchor data is replaced with the \textit{mixup} method explained in Section~\ref{sec:extension}. 
    % Supervised models are trained with standard FL procedure without pre-training on anchor data.
    %\xq {provisional table for wide-resnet results, everything is at 500 rounds. noniid need to run for longer, but we want to show that our method converge faster, and to be fair, we show everything at round 500. maybe we can say that svhn the baseline is high enough, so that gaining extra would be difficult.}
    % ``Pre-trained" represents the models trained with anchor data in prior to FL. 
    
    \vspace{1mm}
    \label{tab:wideresnet}
    \centering
    \renewcommand{\arraystretch}{1.5}
    \scalebox{0.675}{
    \begin{tabular}{lc ccc cc cc}
    \toprule
    \multicolumn{2}{l}{Datasets} & \multicolumn{3}{c}{CIFAR10} & \multicolumn{2}{c}{CIFAR100} & \multicolumn{2}{c}{SVHN} \\
    \hline
    \multicolumn{2}{l}{Number of anchor data} & 250 & 500 & 5000 & 2500 & 10000 & 250 & 1000 \\
    \hline
    % \multicolumn{2}{l}{Pre-trained} &  &  &  &  &  &  &  \\
    % \hline
    \multirow{6}{*}{IID ($Dir=1000$)} & Supervised & 83.91 ± 1.82 & 84.48 ± 0.48 & 80.81 ± 1.05 & 61.15 ± 0.69  & 63.38 ± 0.19 & 92.86 ± 1.15 & 91.41 ± 0.70 \\
    \cline{2-9}
    & FedAvg+FixMatch & 46.59 ± 1.56 & 54.50 ± 1.60  & 77.57 ± 1.17  & 32.71 ± 0.19  & 50.99 ± 0.28 & 60.87 ± 1.49 & 83.07 ± 2.16 \\    
    & FedCon & 39.43 ± 1.08 & 53.25 ± 1.28 & 75.82 ± 1.54 & 30.82  ± 0.19 & 49.04 ± 0.86  & 57.08 ± 0.73  & 81.57 ± 1.49\\
    & SemiFL & 71.83 ± 1.10 & 67.29 ± 2.98  &  77.04 ± 0.50 &  38.86 ± 0.33 & 52.50 ± 0.23 & 90.36 ± 0.43 & 89.39 ± 0.39  \\
    & FedAnchor &  79.52 ± 0.07 & \textbf{81.73 ± 0.43} & 80.72 ± 0.41 &  44.98 ± 0.30 & 53.56 ± 0.45  & \textbf{90.85 ± 0.27} & 91.86 ± 0.09\\
    & FedAnchor (\textit{mix}) & \textbf{80.06 ± 0.62} &  81.16 ± 0.43 & \textbf{83.76 ± 0.23} & \textbf{46.79 ± 0.32} & \textbf{56.40 ± 0.11} & 90.18 ± 0.36  & \textbf{91.88 ± 0.18}\\
    \hline
    \hline
    \multirow{6}{*}{Non-IID ($Dir=0.1$)} & Supervised & 78.54 ± 0.52 & 74.85 ± 0.82  & 79.50 ± 0.46 & 55.87 ± 0.63 & 59.01 ± 0.30 & 87.34 ± 0.88 & 87.48 ± 1.49\\
    \cline{2-9}
    & FedAvg+FixMatch &  41.09 ± 1.42 &  53.29 ± 1.20 &  77.32 ± 0.76 & 32.01 ± 0.25  & 50.75 ± 0.47 & 52.96 ± 0.75 & 83.39 ± 0.67\\    
    & FedCon & 41.11 ± 0.31 & 52.25 ± 1.90 & 75.20 ± 0.42 & 30.27 ± 0.98 & 49.16 ± 0.80 & 57.98 ± 1.49  & 76.78 ± 2.57 \\
    & SemiFL &  49.97 ± 0.64 & 56.99 ± 1.85 & 76.86 ± 0.82 & 38.26 ± 0.32 & 52.08 ± 0.31 & 61.87 ± 1.05 & 85.97 ± 0.52 \\
    & FedAnchor &  52.93 ± 0.95 & 63.38 ± 0.64 &  80.33 ± 0.19 & 40.89 ± 0.23 & \textbf{52.85 ± 0.20} & \textbf{63.81 ± 0.41} & \textbf{86.39 ± 0.33} \\
    & FedAnchor (\textit{mix}) & \textbf{55.26 ± 0.92} & \textbf{64.96 ± 0.27} & \textbf{82.70 ± 0.70} & \textbf{42.25 ± 0.38} & 55.26 ± 0.37 & 48.56 ± 0.30 & 86.09 ± 0.06  \\
    \bottomrule
   \end{tabular}  
   }
\end{table*}

\begin{table}[ht!]
    \caption{\small (upper) Comparison of \method with baselines on model Wide-ResNet28x2 under different augmentation setups.
    (lower) Ablation study on model ResNet-18. FedAnchor(w/o contr loss) represents using embeddings similarities for pseudo-labeling without contrastive loss. Both are conducted on Non-IID ($Dir=0.1$) FL setting.
    }
    \vspace{1mm}
    \label{tab:diff_mix}
    \centering
    \renewcommand{\arraystretch}{1.5}
    \scalebox{0.675}{
    \begin{tabular}{l c c c}
    \toprule
    \multicolumn{2}{l}{Datasets} & \multicolumn{1}{c}{CIFAR10} & \multicolumn{1}{c}{SVHN} \\
    \hline
    \multicolumn{2}{l}{Anchor Size} & 500 & 1000  \\
    \hline
    \multirow{7}{*}{\makecell{ Wide \\ ResNet28x2 }}& FedAvg+FixMatch &   53.29 ± 1.20 & 83.39 ± 0.67 \\    
    & SemiFL &  56.99 ± 1.85 &  85.97 ± 0.52   \\
    & SemiFL+no mixup & 62.93 ± 0.85  & 84.93 ± 0.99\\
    & FedAnchor(client mixup) & 63.38 ± 0.64  & 86.39 ± 0.33 \\
    & FedAnchor(no mixup) & 63.15 ± 0.15 & \textbf{87.78 ± 0.31}\\
    & FedAnchor(client \& server mixup ) & 64.96 ± 0.27 & 86.96 ± 0.08\\
    & FedAnchor(server mixup) & \textbf{66.07 ± 0.50} & 86.09 ± 0.06\\
    \hline
    \multirow{3}{*}{\makecell{ ResNet-18 }} & SemiFL &  68.96 ± 0.98  & 87.21 ± 1.66   \\    
    & FedAnchor(w/o contr loss) & 70.90 ± 0.03  & 88.13 ± 0.35 \\
    & FedAnchor & \textbf{72.75 ± 0.63} & \textbf{90.20 ± 0.56} \\
    \bottomrule
   \end{tabular}  
   }
\end{table}

\subsection{Experimental setup} \label{sec:exp_setup}
The information regarding the exact implementation and packages can be found in the Appendix \ref{app:impl}. 

\textbf{Federated datasets}. We conduct experiments on CIFAR-10/100~\citep{krizhevsky2009learning} and SVHN~\citep{netzer2011reading} datasets. The training set is randomly split into labeled anchor data and unlabeled clients' data for all datasets so that the testing set remains the same for all anchor settings. To make a fair comparison, we set the number of labeled anchor data samples for CIFAR-10/100 and SVHN datasets to be \{$250$, $500$, $5000$\}, \{$2500$, $10000$\} and \{$250$, $1000$\} respectively, according to popular SSL setups~\citep{fixmatch,mixmatch}.
To simulate a realistic \textit{cross-device} FL environment using the rest of the data, we generate IID/non-IID versions of datasets based on actual class labels using Latent Dirichlet Allocation (LDA) with coefficient $Dir$~\citep{qiu2022zerofl, reddi2020adaptive}, where a lower value indicates greater heterogeneity. As a result, the datasets are randomly partitioned into $100$ shards with $Dir=\{1000, 0.1\}$ for IID and non-IID settings, respectively. 

\textbf{Training hyper-parameters}. Following some previous literature, such as MixMatch \citep{mixmatch}, FixMatch \citep{fixmatch} or SemiFL \citep{semifl}, we implemented Wide ResNet28x2~\citep{zagoruyko2016wide} as the backbone model for all datasets.
In addition, we also implemented the most standard version of ResNet-18 \citep{he2016deep} to demonstrate the effectiveness of our method on the standard architecture.
The \textit{anchor head} is set to be a linear layer with $128$ dimensions.
During each FL round, $10$ clients are randomly selected to participate in the training for $5$ local epochs.
The FL training lasts for $500$ rounds. Our implementation of FedCon \citep{long2021fedcon} is based on the original GitHub repository (Appendix \ref{app:impl}),
% \footnote{\url{https://github.com/zewei-long/fedcon-pytorch}} 
from which we extracted both the client and server training pipeline and put them in our codebase.
In the original FedCon paper, only the simple model architectures are tested.
We replaced the original backbone with ours to perform a fair comparison.
More details regarding hyperparameters and baseline implementations can be found in Appendix \ref{app:hyperparameter}.

\subsection{Results}

\begin{figure*}[t]
    \centering
    \subfigure[Anchor 250]{\includegraphics[width=0.24\textwidth]{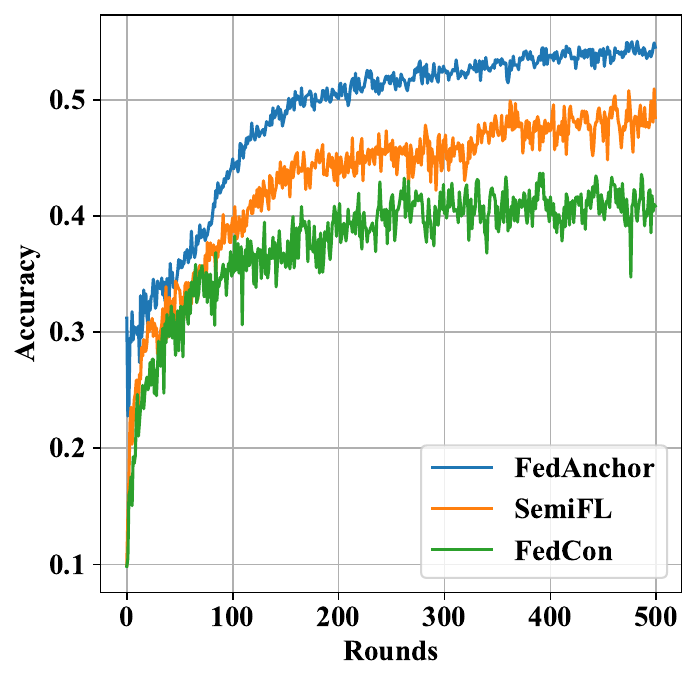}} 
    \subfigure[Anchor 500]{\includegraphics[width=0.24\textwidth]{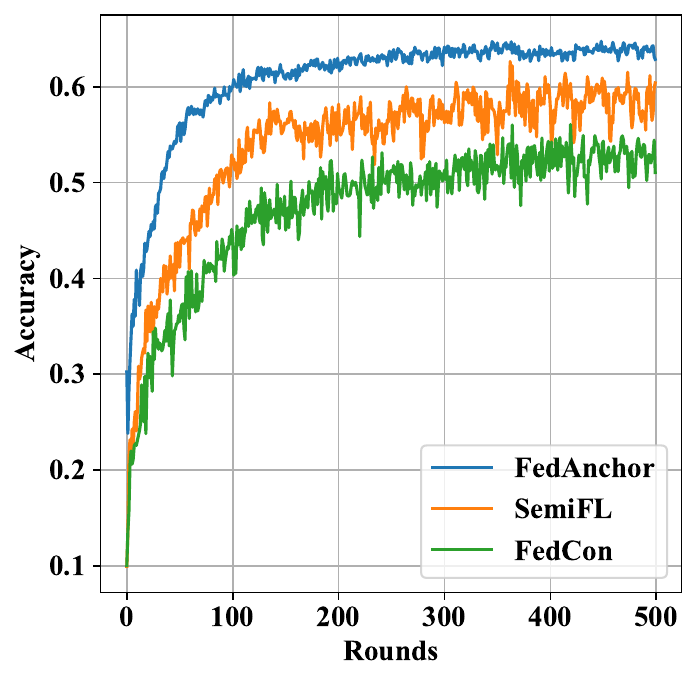}} 
    \subfigure[No. of samples]{\includegraphics[width=0.24\textwidth]{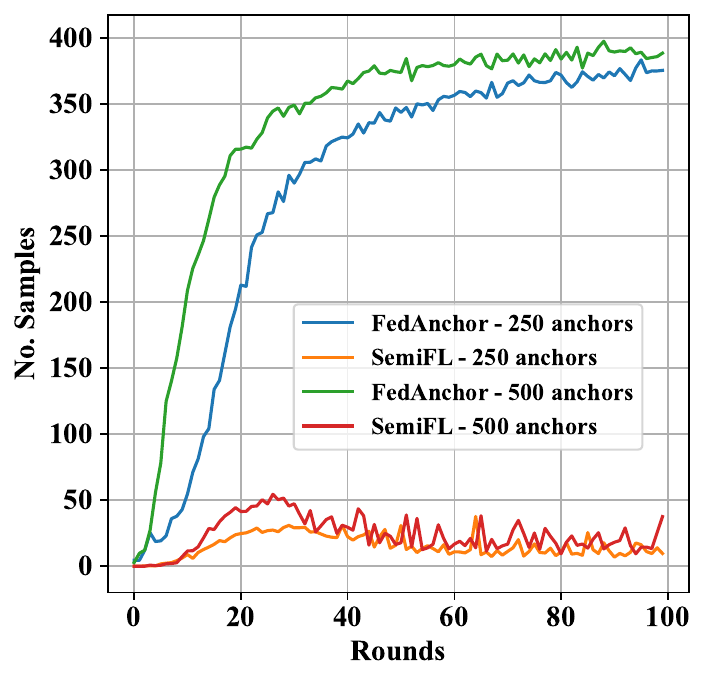}} 
    \subfigure[Pseudo-label Acc]{\includegraphics[width=0.24\textwidth]{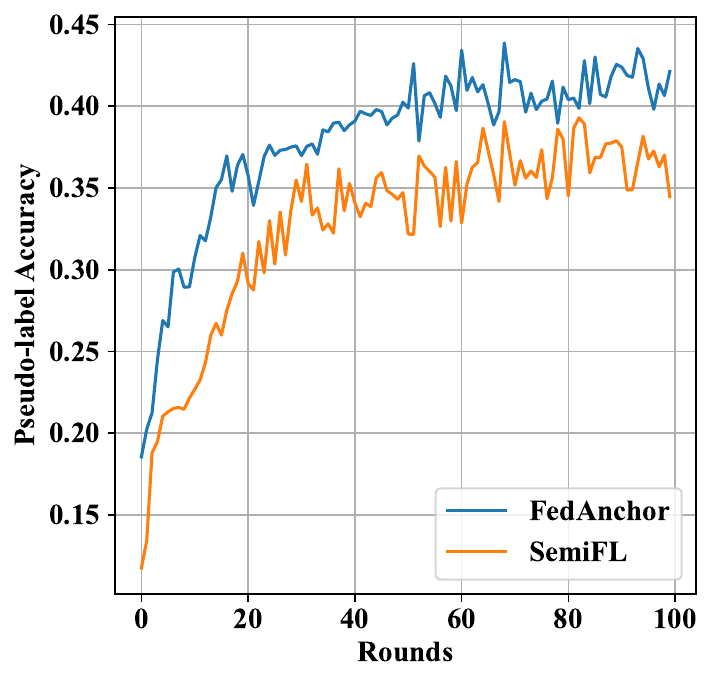}} 
\vspace{-2mm}
\caption{\small Results on CIFAR10 Non-IID with different anchor sizes. (a) and (b) show the testing accuracy with different methods on the two anchor setups. (c) shows the average number of samples selected for training on each client. (d) shows the pseudo-label accuracy of the 500 anchor case (to avoid too many lines in one graph, we show the 500 anchors case, but the general trend is the same for other cases),
which shows that pseudo-labels generated by \method are consistently above the baseline method. (c) and (d) are smoothed every $5$ rounds.}
\label{fig:res_4_plot}
\end{figure*}

Table~\ref{tab:main} and \ref{tab:wideresnet} show the performance of \method along with the baselines Fed+FixMatch, SemiFL~\citep{semifl}, FedCon~\citep{long2021fedcon}. First, the \method outperforms these baselines in all settings and datasets. Specifically, training with a minimal number of anchor data samples (e.g., 25 samples per label) can yield satisfactory performance in IID FL settings. Increasing the anchor data can drastically boost performance in the more challenging but realistic non-IID settings. Indeed, this condition is not difficult to meet as numerous labeled data, suitable to be used as anchors, are stored in centralized data centers. More detailed results can be found in Fig.~\ref{fig:res_4_plot} (a) \& (b), which demonstrates the superiority of \method. Additional graphs can be found in Appendix \ref{app:acc}.

Compared to the baseline SemiFL and FedCon, our methods provide more stable performance, lower standard deviation, and a much faster convergence rate. Indeed, this proves beneficial when deploying this method in real-world applications or industrial contexts. The experimental results for SemiFL under SVHN non-IID with 250 anchor data are extremely volatile and slow to converge. It largely depends on the random process, yielding much unstable accuracy. FedCon consistently underperforms in comparison to SemiFL.

In addition, using the \textit{mixup} method in the server training process obtains slightly enhanced performance in most cases. This improvement is primarily attributable to the ample use of data, which is advantageous for training a more robust model \citep{mixmatch}. Utilizing \textit{mixup} method locally on the client-side training can also boost the performance. %This can also explain why some \method results can be higher than the fully supervised performance, where no \textit{mixup} is implemented. 
%Another reason that \method might achieve higher performance can result from the alternate server training process. \xq{depending on the res of supervised, might modify here.} 
We can observe across all datasets that alternative training with more anchor data samples on the server can make the difference between IID and non-IID smaller because the anchor data can be seen as a ``shared IID data'' that can reduce heterogeneity across clients.

\begin{figure}[!t]
    \centering
    \subfigure[Test Accuracy]{\includegraphics[width=0.23\textwidth]{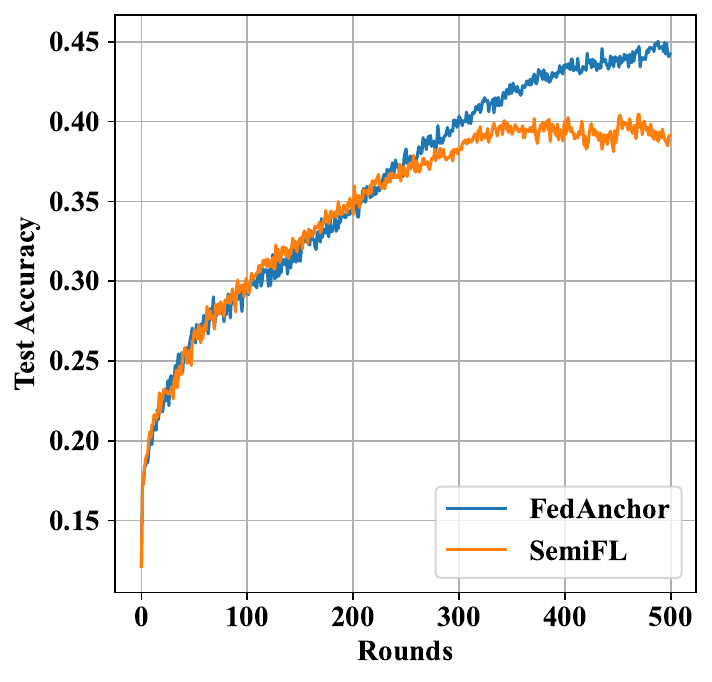}} 
    \subfigure[Pseudo-label Accuracy]{\includegraphics[width=0.23\textwidth]{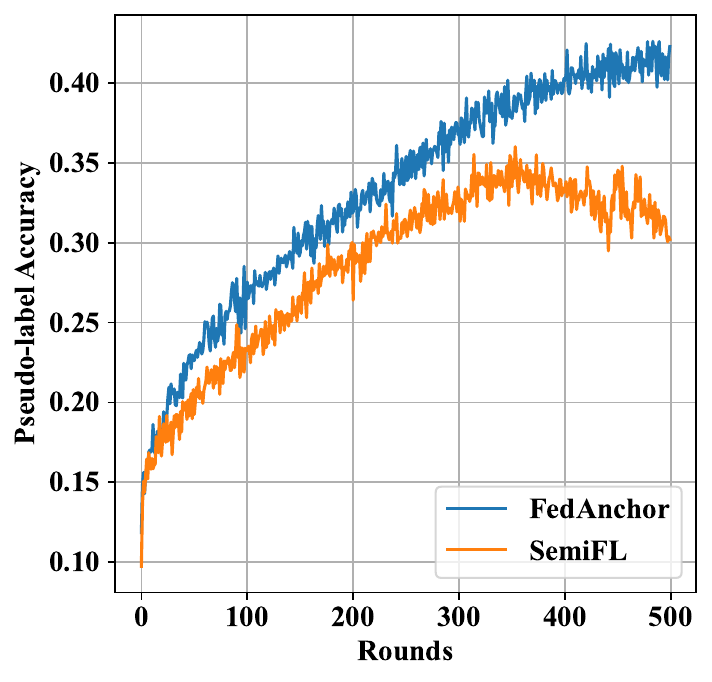}} 
\caption{\small Plots on CIFAR100 (IID, 2500 anchors) experiment. (a) testing accuracy. (b) pseudo-label accuracy. \method can reduce the confirmation bias compared to SemiFL.}
\label{fig:confirmation_bias}
\end{figure}

%% to show the confirmation bias
Figure \ref{fig:confirmation_bias} shows that \method can reduce the confirmation bias compared with the baseline. The figure demonstrates that SemiFL, which uses model prediction as the pseudo-labels, degenerates after $300$ rounds, even with the IID partitions. It is clear from the figure that the accuracy stops increasing, which indicates that the model is overfitting to the easier data samples and generating poorer pseudo labels, resulting in lower testing accuracy. 

We also conduct ablation studies on the naive implementation of FixMatch~\citep{fixmatch}, the centralized SOTA method, under the FL setups, as shown in Table \ref{tab:main} \& \ref{tab:wideresnet}, to show the effect of the label contrastive loss and the proposed pseudo-labeling. As mentioned in Section \ref{sec:related}, the original centralized setting of FixMatch requires sampling both labeled and unlabeled data per mini-batch. Therefore, implementing FedAvg+FixMatch combines FixMatch locally on the client side and alternative training using anchor data on the server side. We can see that the simple combination of FedAvg and FixMatch fails to produce excellent performance.

%explain that diff dataset might want different mixup
In addition, we experiment with different augmentation arrangements as shown in Table \ref{tab:diff_mix}, which demonstrates that utilizing the mixup method on the client side, as SemiFL did, might not always lead to the best performance. However, \method still outperforms in all augmentation cases.

\subsection{Pseudo labeling quality}
The quality of pseudo-labels can largely determine the convergence rate of the training and its resulting performance. As unlabeled client training data is trained with generated hard pseudo-labels, with higher pseudo-label accuracy, the model can extract more useful information from the unlabeled client data, converging to better performance with a faster convergence rate.

As previously shown in Section \ref{sec:keyobservations}, the average number of samples trained by each client for the method SemiFL can be less than $10 \%$ of the total samples stored on the client side. We compare the average number of samples trained by clients for \method with SemiFL in Fig.~\ref{fig:res_4_plot}(c). As demonstrated in the figure, the average number of samples trained by each client grows with the training round, which is as expected since the model is trained to improve gradually. The average number is growing to almost $400$ for \method compared with consistently below $100$ for SemiFL. This is presumably attributed to the higher quality of the pseudo-labels generated by \method.

To substantiate this claim, we compute the pseudo-label accuracy by comparing the pseudo-labels generated by different methods with the true labels, as shown in Fig.~\ref{fig:res_4_plot} (d) \& \ref{fig:confirmation_bias}(b). Clearly, \method can produce higher pseudo-label accuracy with a big margin compared to the baseline method.
More results can be found in Appendix \ref{app:plabel_quality}.
% In addition, we also measure the quality of pseudo-labeling in two different aspects. We compute the pseudo-label accuracy by comparing the pseudo-labels generated by different methods with the true labels, as shown in Figure \ref{fig:res_4_plot} (d) and \ref{fig:confirmation_bias}(b). Fig.~\ref{fig:pip} (right) shows the pseudo-label accuracy at the round number 100, demonstrating that the \method can produce higher pseudo-label accuracy with a big margin. More results can be found in Appendix \ref{app:plabel_quality}.

\subsection{Ablation studies}
We conduct an ablation study by refining the pseudo-labeling method in SemiFL using the similarities between feature embeddings of unlabeled data and labeled anchor data, notably \textit{excluding the use of contrastive loss} in the training process. One can see from Table \ref{tab:diff_mix} (lower part) that adopting our new pseudo-labeling approach with anchor data slightly boosts the performance. 
However, including contrastive loss is imperative to achieve a significantly higher accuracy.

\subsection{Communication Overhead}\label{app:overhead}
As we know from \cite{qiu2023first}, communication cost is one of a big concern in FL. Table \ref{tab:communicationoverhead} shows the communication overhead of \method compared to standard supervised FL training. The overhead is \textit{only for downstream communication} when extra anchor embeddings must be sent from the central server to the selected clients. The overhead is calculated as the percentage of transmitted parameters that compose the anchors' embeddings over the number of the model's parameters transmitted by \emph{FedAvg}: $Overhead = \frac{100}{\dim(\Bar{w})}\sum_{i=1}^{S}\dim(z_i^{anchor})$, where $\Bar{w}$ is the model transmitted by \emph{FedAvg}. The upstream communication is the same as \emph{FedAvg}. Thus, the extra communication overhead of \method is negligible compared to the one during standard FL training.

\begin{table}[t]
    \caption{\small Downstream communication overhead of \method compared with supervised FL training. 
    }
    % \vspace{3mm}
    \label{tab:communicationoverhead}
    \centering
    \scalebox{0.8}{
    \begin{tabular}{l cc}
    \toprule
    Datasets & Anchor Size & Overhead \\
    \midrule
    \multirow{1}{*}{CIFAR10}  & 250 / 500 / 5000&  0.29$\%$ / 0.57$\%$ / 5.73$\%$ \\
    %& 500 & 0.57$\%$ \\
    %& 5000 & 5.73$\%$ \\
    %\midrule
    \multirow{1}{*}{CIFAR100}  & 2500 / 10000 & 2.85$\%$ / 11.41$\%$\\
    %& 10000 & 11.41$\%$ \\
    \multirow{1}{*}{SVHN}  & 250 / 1000 & 0.29$\%$ / 1.15$\%$  \\
    %& 1000 & 1.15$\%$ \\  
    \bottomrule
    \end{tabular}
}
\end{table}
\section{Conclusion}\label{sec:conclusion}

In this paper, we propose \method, which is a FSSL method enhanced by a newly designed label contrastive loss based on the cosine similarity to train on labeled anchor data on the server.
Instead of retaining the high-confidence data solely through model predictions as in the conventional SSL studies, \method generates the pseudo labels by comparing the similarities between the model representations of unlabeled data and labeled anchor data. This provides better quality pseudo-labels, alleviates the confirmation bias, and reduces the issues of over-fitting to easy-to-learn data samples. We perform extensive experiments on three different datasets with different sizes of labeled anchor data on the server and show that the proposed methods achieve state-of-the-art performance with a faster convergence rate. As for future direction, we are experimenting with a fixed threshold during the pseudo-labeling using the anchor data stage. More advanced adaptive and dynamic thresholding techniques can be implemented to improve the performance and convergence rate further. 

\section*{Impact Statement}
This paper presents work whose goal is to advance the field of Decentralized Machine Learning. There are many potential societal consequences of our work, none of which we feel must be specifically highlighted here.

\bibliography{example_paper}
\bibliographystyle{icml2024}

%%%%%%%%%%%%%%%%%%%%%%%%%%%%%%%%%%%%%%%%%%%%%%%%%%%%%%%%%%%%%%%%%%%%%%%%%%%%%%%
%%%%%%%%%%%%%%%%%%%%%%%%%%%%%%%%%%%%%%%%%%%%%%%%%%%%%%%%%%%%%%%%%%%%%%%%%%%%%%%
% APPENDIX
%%%%%%%%%%%%%%%%%%%%%%%%%%%%%%%%%%%%%%%%%%%%%%%%%%%%%%%%%%%%%%%%%%%%%%%%%%%%%%%
%%%%%%%%%%%%%%%%%%%%%%%%%%%%%%%%%%%%%%%%%%%%%%%%%%%%%%%%%%%%%%%%%%%%%%%%%%%%%%%
\newpage
\appendix
\onecolumn
\section{Algorithm pseudo code}
\begin{algorithm}[!h]
    %\small
    \caption{\small\textbf{\method}: Let $N$ be the total number of clients, with total $n$ data samples; $(x^m)$ be the unlabeled data in client $m$; $(x^{anchor}, y^{anchor})$ be the labeled anchor data in the server; $z^m$ be the output of the latent space for the unlabeled client data, and $z^{anchor}$ be the output of the latent space for the anchor data; we let $\hat{y}$ be the pseudo-label for unlabeled data; $E$ be the number of local epochs; $r$ be the participation ratio of clients in each round; $w_t$ be the aggregated weights at round $t$; $C$ be the total number of labels; $Z^{anchor}$ be the set of anchor latent representation; $\eta_{clt}$ and $\eta_{ser}$ be the client and server learning rate respectively. \textbf{ClientTraining} be the training procedure on the client using the local dataset, and \textbf{ServerTraining} be the training procedure on the server using the anchor dataset.} \label{algorithm:algo}
    \begin{algorithmic}[1]
        %\Procedure{Server executes}{\  }
        \STATE \textbf{Procedure} \textbf{Server executes}
            \STATE Initialize the model weight $w_0$
            \FOR {t = 1,...,T}
                \STATE $M_t \leftarrow$ random select $(rN)$ number of clients 
                \STATE compute $Z^{anchor}$
                \FOR{ each $m \in M_t$}
                    \STATE $\hat{y}^m \leftarrow$ Pseudo Label($x_m, Z^{anchor}$) 
                    \STATE $w_{t+1}^m \leftarrow$  ClientTraining$(w_t, x_{m},\hat{y}^m)$
                \ENDFOR
            \STATE $w_{t+1} \leftarrow \sum_{m \in M_t} \frac{n_m}{\sum n_m} w_{t+1}^m$
            \STATE $w_{t+1} \leftarrow$ ServerTraining($w_{t+1}, x^{anchor}, y^{anchor}$)
        \ENDFOR
        %\EndProcedure

    %\Procedure{ClientUpdate}{$k,w_t,x_{m},Z_a$}
    %    \State $\hat{y}_m \leftarrow$ PseudoLabel($x_{m}, Z_a$)
    %    \State create fix dataset: $\mathcal{D}^{fix}_m = \{  (x_{m, i}, \hat{y}_{m, i} )$ s.t. $\max(sa(z_{m, i})_c) > t \}$ (eq. \ref{eq:fixdataset})
    %    \If{ $\mathcal{D}^{fix}_m \not = \emptyset$}
    %        \State create mix dataset: $\mathcal{D}^{mix}_{m} = $Sample $|\mathcal{D}^{fix}_{m}|$  with replacement from  $\{ (x_{m, i}, \hat{y}_{m, i} )\}$ (eq. \ref{eq:mixdataset})
    %    \Else
    %        \State terminate ClientUpdate
    %    \EndIf
    %    \For{local epoch $e = 1,..., E$}
    %    \State $\mathcal{L}_{fix} = l(f(A(x^{fix}_b), \hat{y}^{fix}_b)$ (eq. \ref{eq:fixloss})
    %    \State $\lambda_{mix} \sim Beta(a,a), \ x_{mix} = \lambda_{mix} x^{fix} + (1-\lambda_{mix}) x^{mix}$  (eq. \ref{eq:mixloss1})
    %    \State $\mathcal{L}_{mix} = \lambda_{mix} \cdot l(f(\alpha(x_{mix}, \hat{y}^{fix})) + (1-\lambda_{mix} ) \cdot l(f(\alpha(x_{mix}, \hat{y}^{mix}))$  (eq. \ref{eq:mixloss2})
    %    \State $\mathcal{L}_{combine} = \mathcal{L}_{fix} + b\mathcal{L}_{mix}$  (eq. \ref{eq:combinedloss})
    %    \State $w_{t+1}^m \leftarrow w_t - \eta_{clt} \nabla \mathcal{L}_{combine}$
    %    \EndFor
    %\EndProcedure
    \vspace{2mm}
    
    \STATE \textbf{Procedure} \textbf{Pseudo Label($x_{m}, Z^{anchor}$)}
    %\Procedure{PseudoLabel}{$x_{m}, Z_a$}
    \FOR{i = 1,...,$|x_m|$}
        \STATE create label-specific anchor projected set like below:
        \STATE $Z^{anchor}_{c} = \{ z^{anchor}_{i}, y^{anchor}_i = c \}, \forall c \in [C] $
        \STATE $s^{avg}_c(z^m_{i}) = \frac{\sum_{z_k \in Z^{anchor}_{c}} s(z^m_{i}, z_k)  }{|Z^{anchor}_{c}|}$ (eq. \ref{eq:savg})
        \STATE $\hat{y}^m_{i} = argmax_{c \in [C]} \big( s^{avg}_c(z^m_{i})\big)$  (eq. \ref{eq:yhat})
    \ENDFOR
    %\EndProcedure
    %\Procedure{ServerUpdate}{$w_{t+1}$}
    %    \State compute supervised loss $\mathcal{L}_s = l(f(A(x_a), y_a)$
    %    \State $w_{t+1} \leftarrow w_{t+1} - \eta_{ser}\nabla \mathcal{L}_s$
    %    \State $l(c) = -\log \frac{\sum _{y(i), y(j) = c}\exp (s_{z_i,z_j}/\tau)}{\sum_{y(i) \not = y(j) %} \exp(s_{z_i,z_j}/\tau)}  $(eq. \ref{eq:cossimloss})
    %    \State $\mathcal{L}_c =  \frac{1}{C} \sum_{c=1}^C l(c) $ (eq. \ref{eq:cossimloss2})
    %    \State $w_{t+1} \leftarrow w_{t+1} - \eta_{ser}\nabla \mathcal{L}_c$
    %\EndProcedure
    \end{algorithmic}
\end{algorithm}

\section{Mixup training methods}\label{app:mixup}
In this section, we explain the mixup method implemented in detail.

Each selected client needs to construct a high-confidence dataset $\mathcal{D}^{fix}_{m}$, which is called the \textit{fix dataset} inspired by FixMatch \citep{fixmatch} and SemiFL \citep{semifl}. The \textit{fix dataset} is defined to be the set of data samples with the similarity scores above the preset threshold $t$:
%\vspace{-1mm}
\begin{equation}\label{eq:fixdataset}
    \mathcal{D}^{fix}_m = \{  (x^m_{i}, \hat{y}^m_{i} ) \ \ \verb!s.t.! \max\big(s^{avg
}_c(z^m_{i})\big) > t \}.
\end{equation}

The current local training will be stopped if the client has an empty \textit{fix dataset}. Otherwise, we then will sample with replacement to construct a \textit{mix dataset} inspired by MixMatch \citep{mixmatch} as below:
%\vspace{-1mm}
\begin{equation}\label{eq:mixdataset}
    \mathcal{D}^{mix}_{m} = \verb!Sample! |\mathcal{D}^{fix}_{m}| \verb! with replacement from ! \{ (x^m_{i}, \hat{y}^{m}_{i} )  \}
\end{equation}

where $|\mathcal{D}^{fix}_{m}|$ represents the size of the \textit{fix dataset}, and in this case $|\mathcal{D}^{fix}_{m}| = |\mathcal{D}^{mix}_{m}|.$

During local training with a nonempty \textit{fix dataset}, the loss function $\mathcal{L}$ consists of two parts: $\mathcal{L}_{fix}$ and $\mathcal{L}_{mix}$. $\mathcal{L}_{fix}$ is calculated as in standard supervised training with mini-batch sampled from the \textit{fix dataset}, but attaching strong augmentation $A(\cdot)$ on each data input:
%\vspace{-1mm}
\begin{equation}\label{eq:fixloss}
    \mathcal{L}_{fix} = l\big(f\big(A(x^{fix}_b), \hat{y}^{fix}_b\big)\big). 
\end{equation}
where $l$ is the loss function, such as cross-entropy loss for classification tasks. 

In addition, the \textit{mix loss} $\mathcal{L}_{mix}$ is computed following the Mixup method. Client $m$ constructs a mixup data sample from one fix data $x^{fix}$ and one \textit{mix dataset} $x^{mix}$ by:

\begin{equation}\label{eq:mixloss1}
    \lambda_{mix} \sim Beta(a,a), \ x_{mix} = \lambda_{mix} x^{fix} + (1-\lambda_{mix}) x^{mix} ,
\end{equation}
where $Beta(a,a)$ represents the beta distribution and $a$ is a \textit{mixup hyperparameter}, and the \textit{mix loss} is calculated as:

\begin{equation}\label{eq:mixloss2}
    \mathcal{L}_{mix} = \lambda_{mix} \cdot l\big(f\big(\alpha(x_{mix}, \hat{y}^{fix})\big)\big) + (1-\lambda_{mix} ) \cdot l\big(f\big(\alpha(x_{mix}, \hat{y}^{mix})\big)\big), 
\end{equation}
where $\alpha(\cdot)$ represents weak augmentation of data samples.
A single local epoch of a client $m$ corresponds to applying as many local SGD steps on the combined loss as the number of batches it has in $\mathcal{D}^{mix}_{m}$:
%\vspace{-1mm}
\begin{equation}\label{eq:combinedloss}
    \mathcal{L}_{combine} = \mathcal{L}_{fix} + b\mathcal{L}_{mix},
\end{equation}
where $b$ can be a linear combination coefficient and set to be $1$ as default. 

Finally, after operating for $E$ local epochs, client $m$ returns the updated model parameters to the central server to finish the local training.

\section{Training Hyper-parameters} \label{app:hyperparameter}
Before FL, the model is pre-trained on anchor data on the server for $5$ epochs with SGD to speed up the training process with a learning rate of $0.05$. We use a learning rate of $0.03$, a weight decay of $5e-4$, and an SGD momentum of $0.9$ for both local training on the client side and anchor training on the server side. As indicated in the paper, the threshold is set to $0.6$ for our method and $0.95$ for the SemiFL baseline. We conduct three random experiments for all the datasets with different seeds, and the performance is reported on the centralized test set. The hyperparameter for the beta distribution of \textit{mixup} (eq.\ref{eq:mixloss1}) is $a=0.75$, and the linear coefficient combining \textit{fix loss} and \textit{mix loss} $b$ (eq. \ref{eq:combinedloss}) is $1$ for all experiments. We implement RandAugment~\citep{cubuk2020randaugment} as a robust augmentation method.

We have also included the federated supervised training as one of the baselines. The supervised implementation follows the standard experimental protocol as per previous papers \citep{reddi2020adaptive,qiu2022zerofl,horvath2021fjord}, where no strong augmentation methods are implemented.

\section{Latent space visualization}\label{app:latent_space}
\begin{figure*}[t]
    \centering
    \subfigure[]{\centering\includegraphics[width=0.4\textwidth]{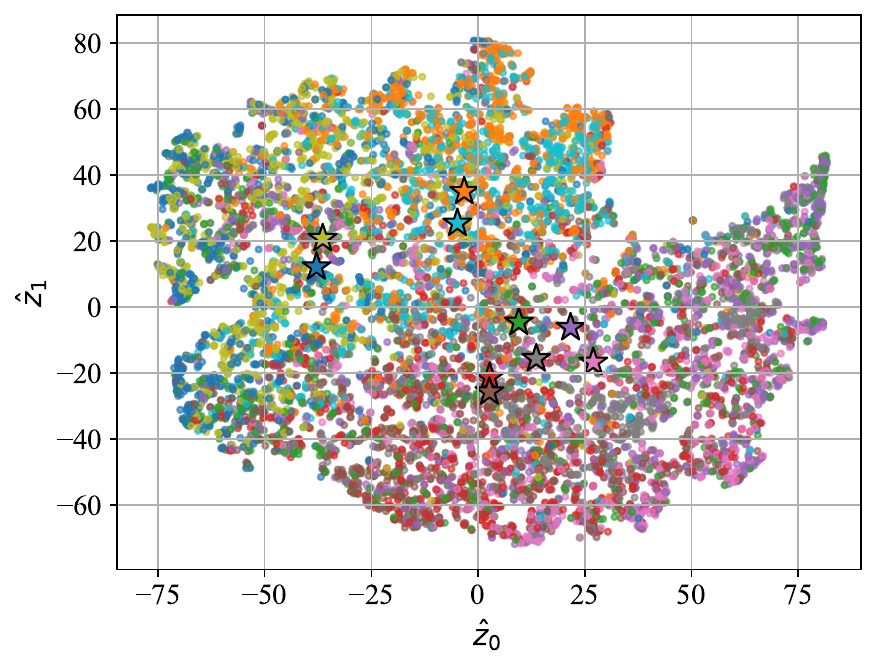}} 
    \subfigure[]{\centering\includegraphics[width=0.4\textwidth]{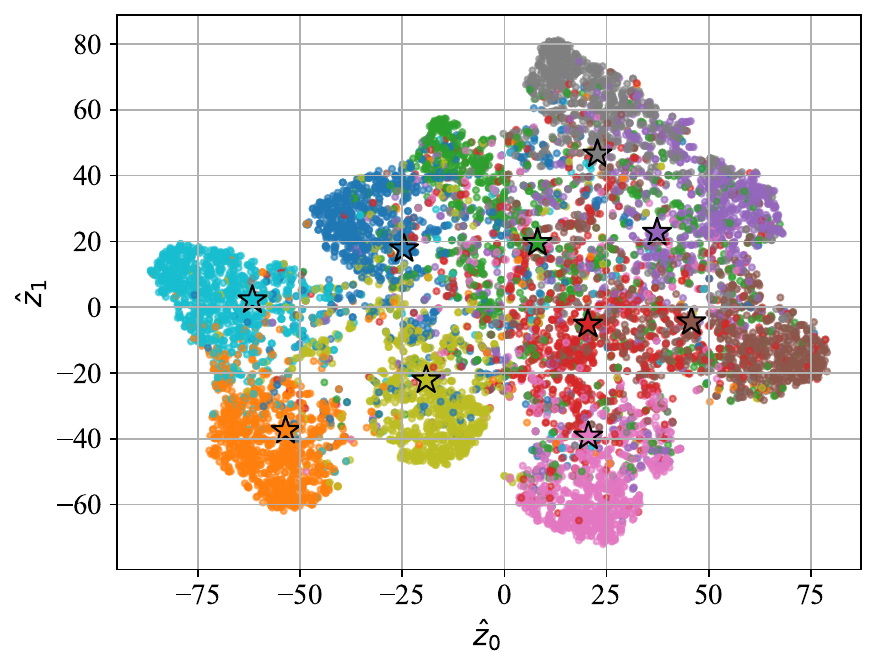}} 
    \subfigure[]{\centering\includegraphics[width=0.4\textwidth]{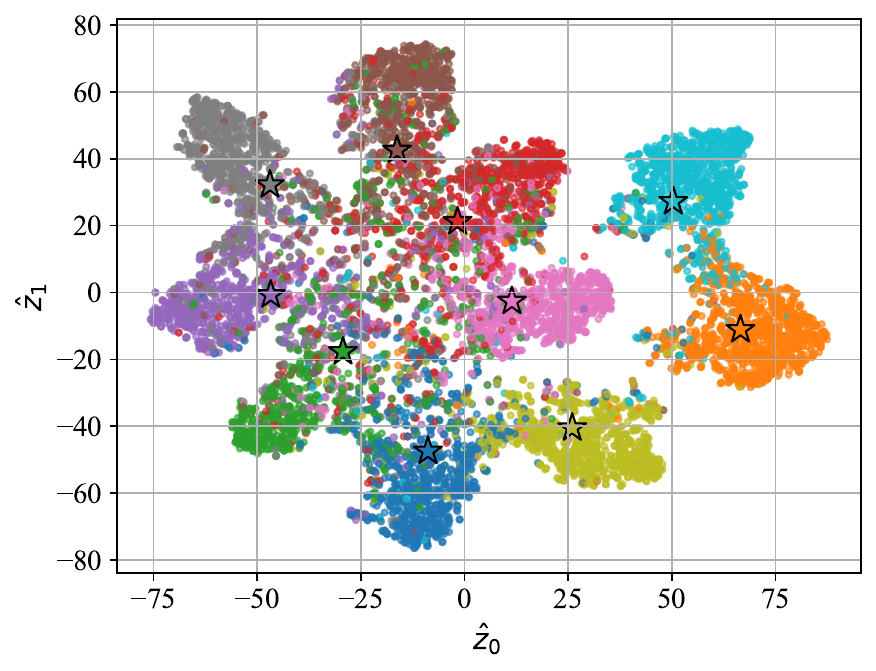}} 
    \subfigure[]{\centering\includegraphics[width=0.4\textwidth]{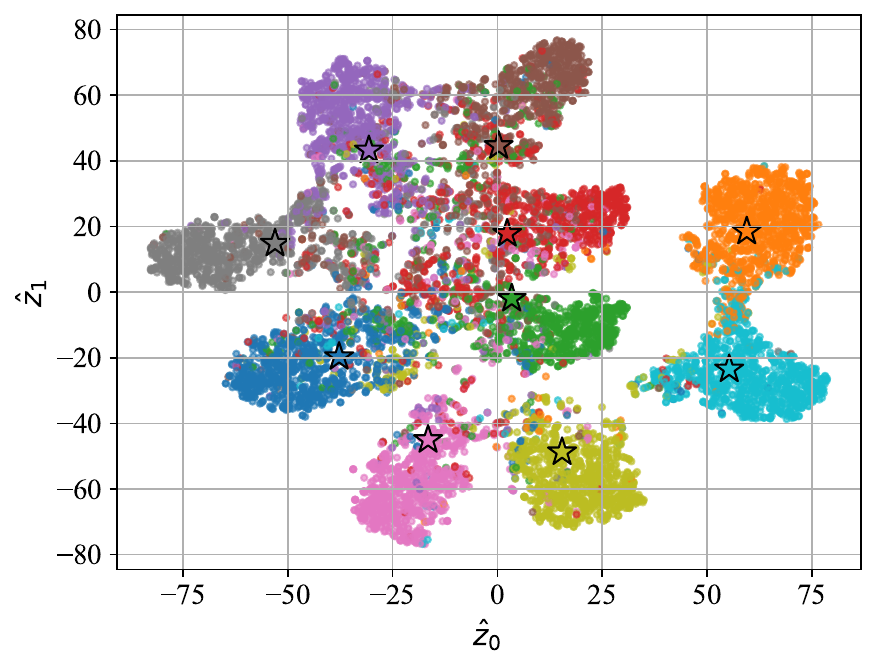}} 

    \caption{\small Visualization of the latent space of the model at different stages of the FL using \method. A t-SNE dimensionality reduction has been performed to improve readability. CIFAR-10 with Wide ResNet backbone is represented here. Different colors refer to different labels. Each data point is representative of a data sample in the (centralized) test set of CIFAR-10. The stars \big(\FiveStar \big) represent the centroids for each label. The rounds represented in (a), (b), (c), and (d) are respectively $1$ (after one aggregation), $100$, $200$, and $500$ (final global model). }
    \label{fig:latent_vis}
\end{figure*}

Figure \ref{fig:latent_vis} provides a visualization of the latent space of the model at different stages of the FL using \method. The visualization demonstrates that \method can successfully separate the classes in the latent space through the training process.

\begin{figure*}[t]
    \centering
    \subfigure[IID ($Dir=1000$) - 250 anchors]{\centering\includegraphics[width=0.31\textwidth]{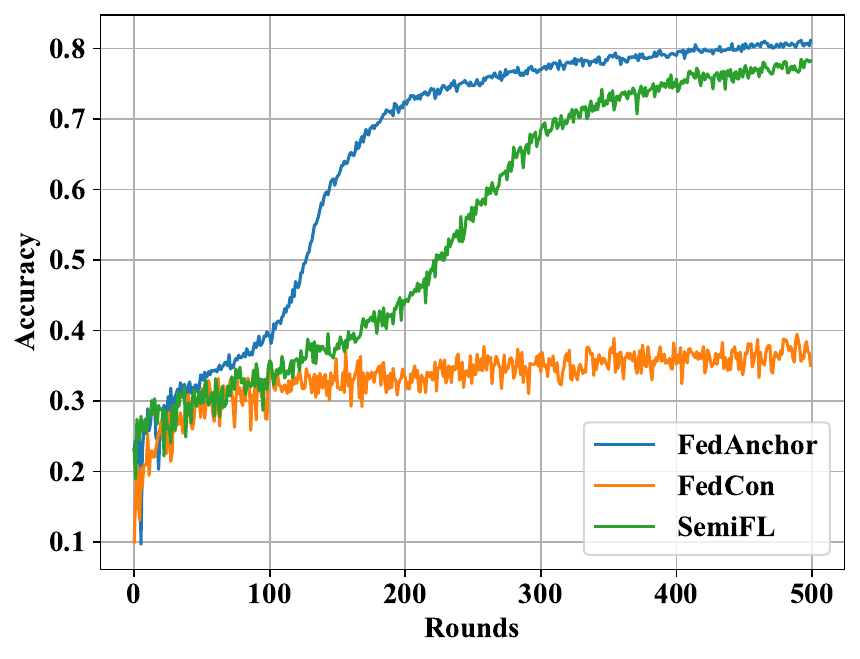}} 
    \subfigure[IID ($Dir=1000$) - 500 anchors]{\centering\includegraphics[width=0.31\textwidth]{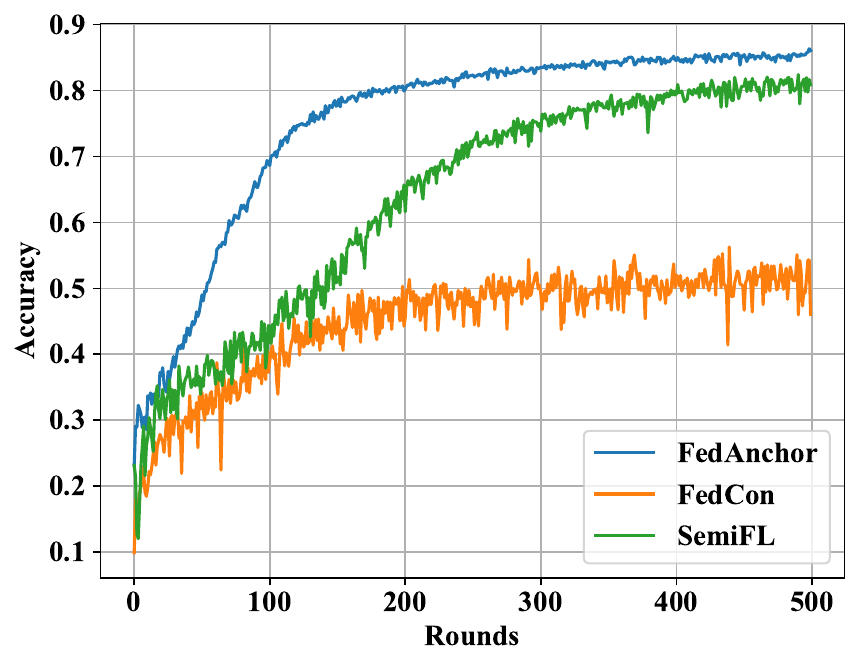}} 
    \subfigure[IID ($Dir=1000$) - 5000 anchors]{\centering\includegraphics[width=0.31\textwidth]{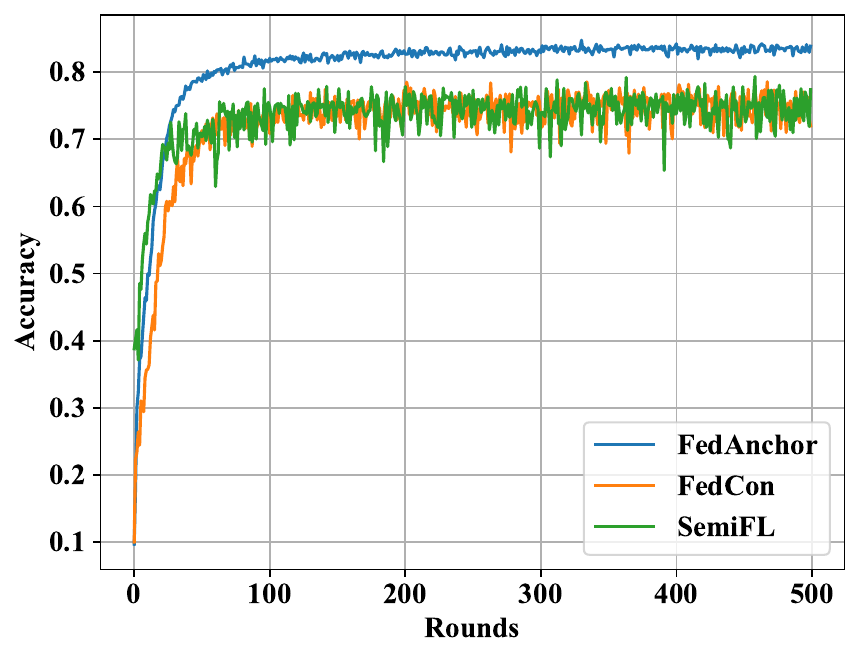}} 
    \subfigure[non-IID ($Dir=0.1$) - 250 anchors]{\centering\includegraphics[width=0.31\textwidth]{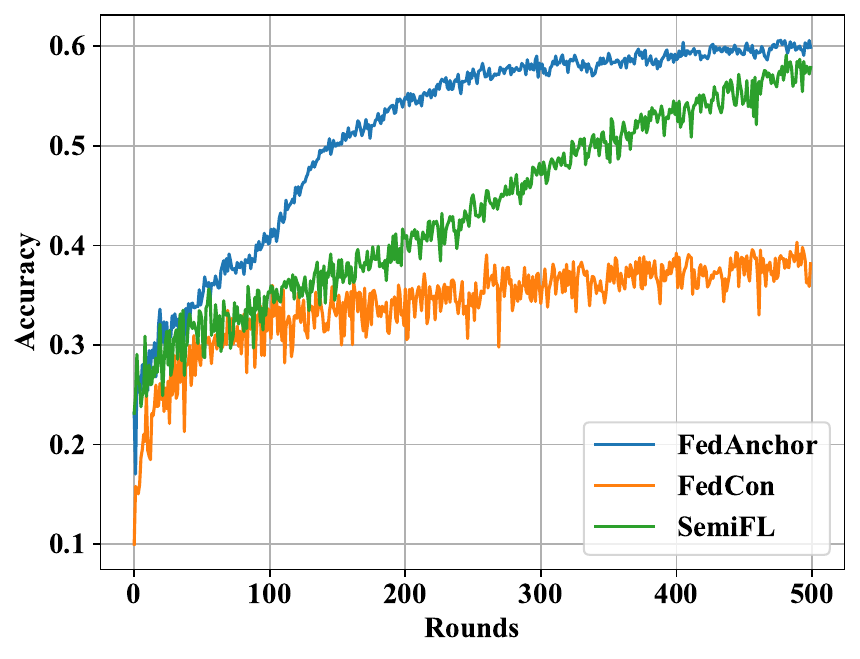}} 
    \subfigure[non-IID ($Dir=0.1$) - 500 anchors]{\centering\includegraphics[width=0.31\textwidth]{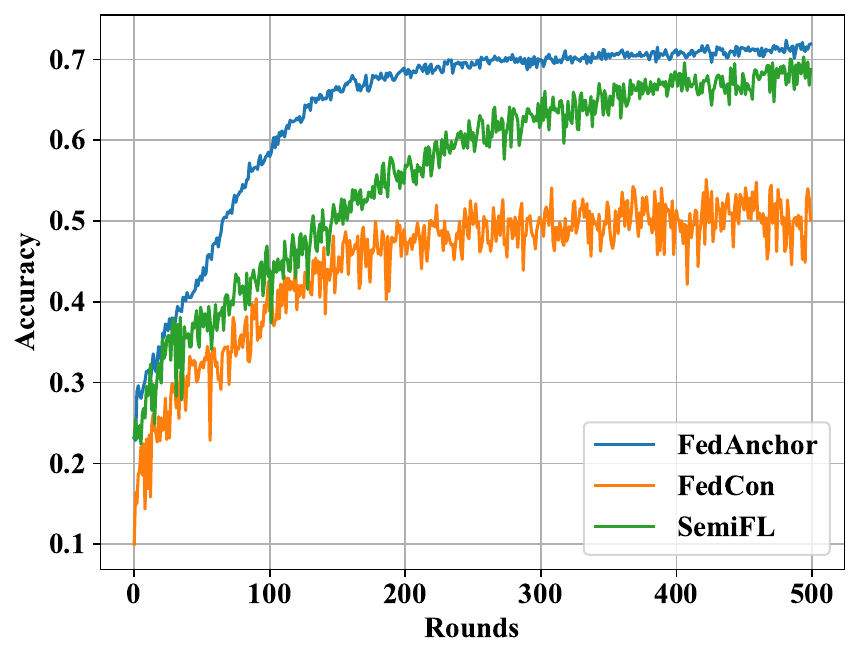}} 
    \subfigure[non-IID ($Dir=0.1$) - 5000 anchors]{\centering\includegraphics[width=0.31\textwidth]{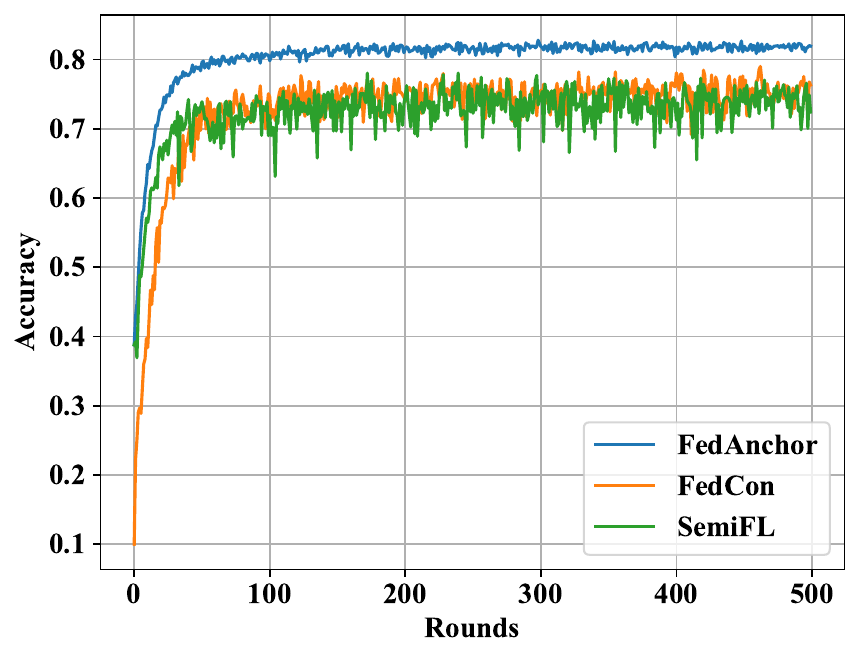}} 
\caption{\small Performance on CIFAR10 test set on ResNet-18 with both IID and non-IID cases across different numbers of anchor data setups, compared with the baseline.}
\label{fig:acc_rounds}
\end{figure*}

\section{More Experimental Results} \label{app:acc}
\begin{figure}[h]
    \centering
    \subfigure[IID ($Dir=1000$) - 2500 anchors]{\centering\includegraphics[width=0.4\textwidth]{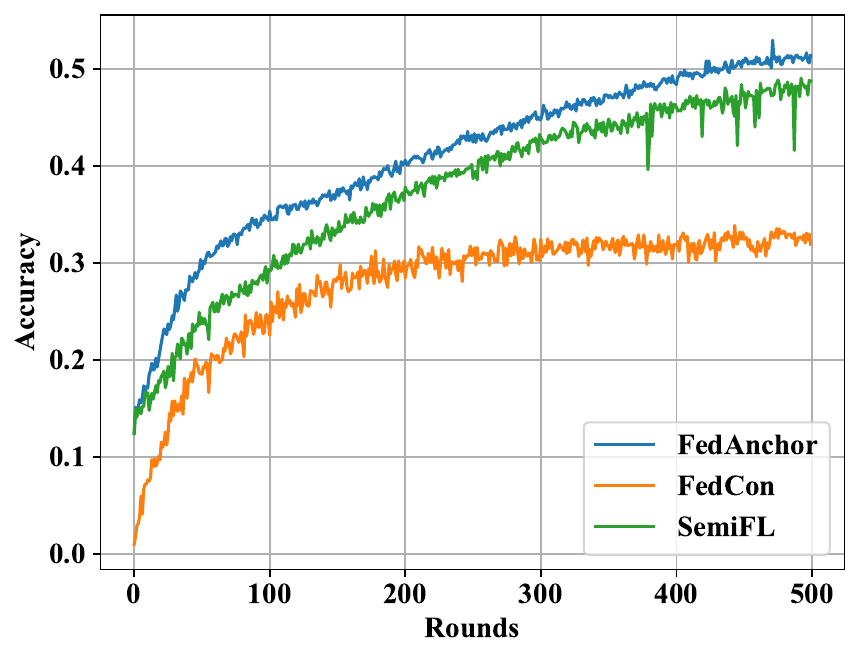}} 
    \subfigure[IID ($Dir=1000$) - 10000 anchors]{\centering\includegraphics[width=0.4\textwidth]{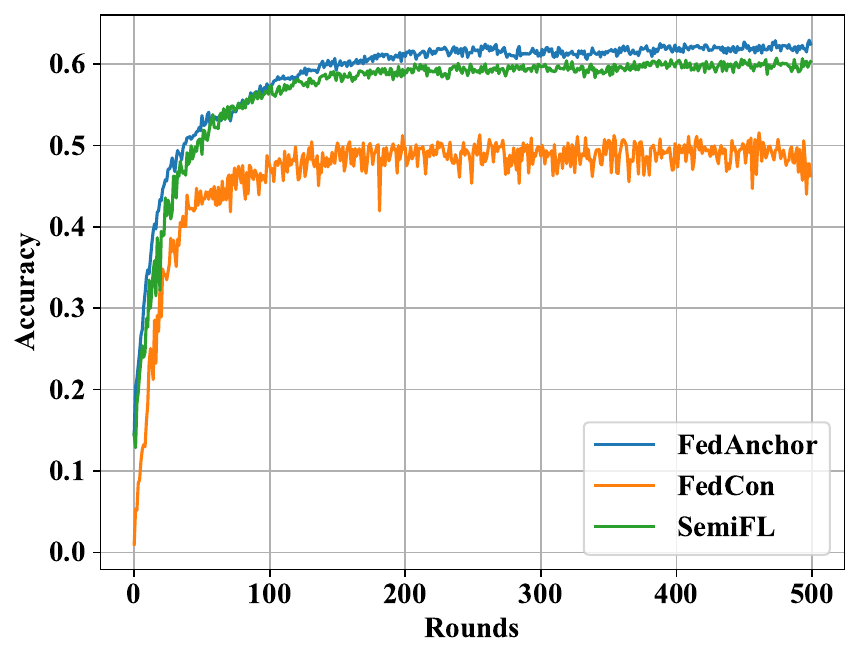}} 
    \subfigure[non-IID ($Dir=0.1$) - 2500 anchors]{\centering\includegraphics[width=0.4\textwidth]{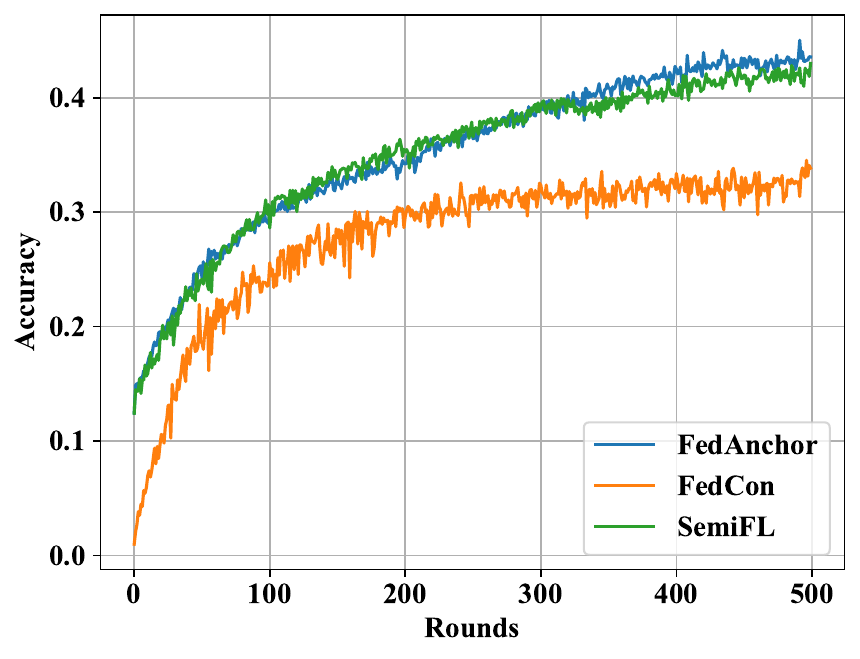}} 
    \subfigure[non-IID ($Dir=0.1$) - 10000 anchors]{\centering\includegraphics[width=0.4\textwidth]{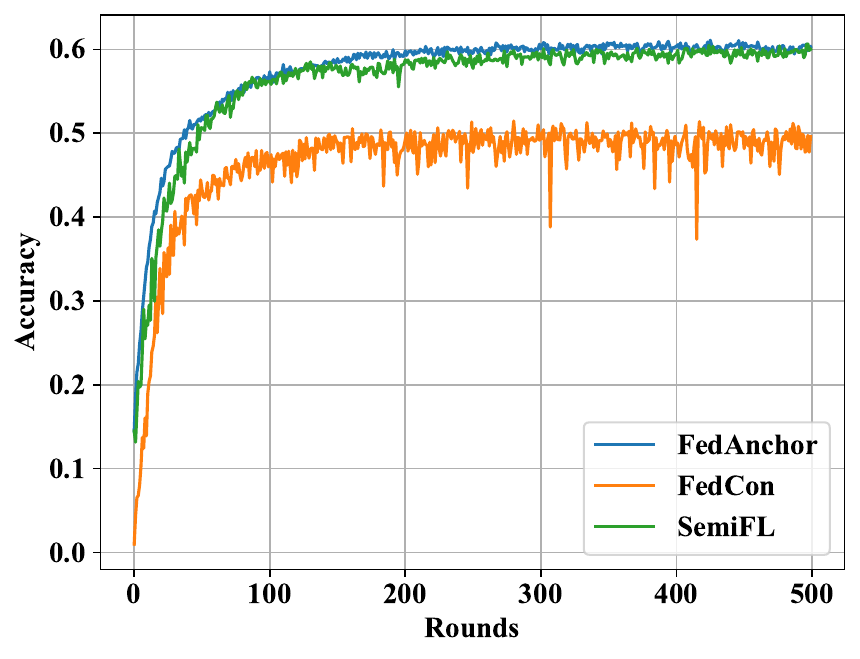}} 
\caption{\small Performance on CIFAR100 test set on ResNet-18 with both IID and non-IID cases across different numbers of anchor data setups, compared with the baseline.}
\label{fig:acc_rounds_cifar100}
\end{figure}

\begin{figure}[h]
    \centering
    \subfigure[IID ($Dir=1000$) - 250 anchors]{\centering\includegraphics[width=0.4\textwidth]{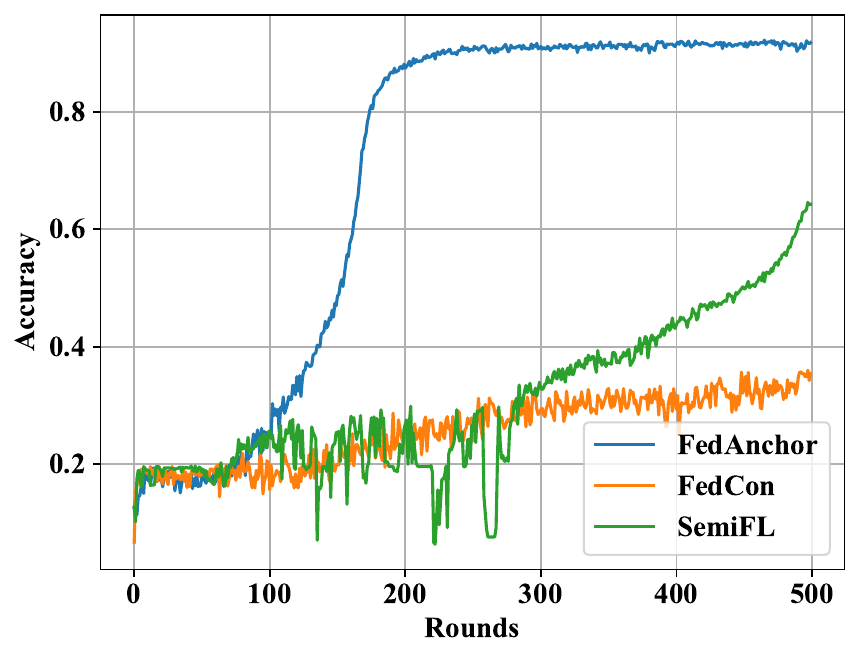}} 
    \subfigure[IID ($Dir=1000$) - 1000 anchors]{\centering\includegraphics[width=0.4\textwidth]{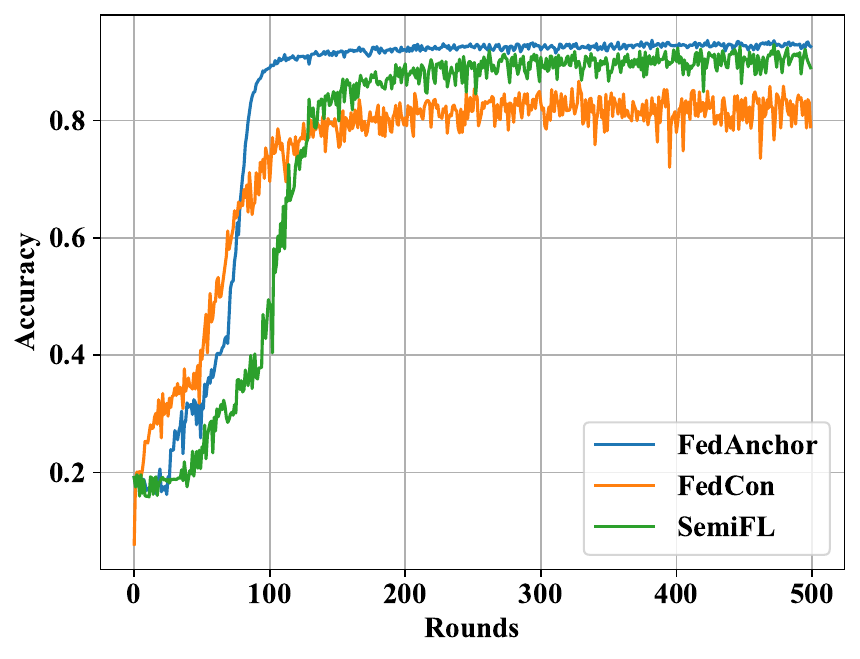}} 
    \subfigure[non-IID ($Dir=0.1$) - 250 anchors]{\centering\includegraphics[width=0.4\textwidth]{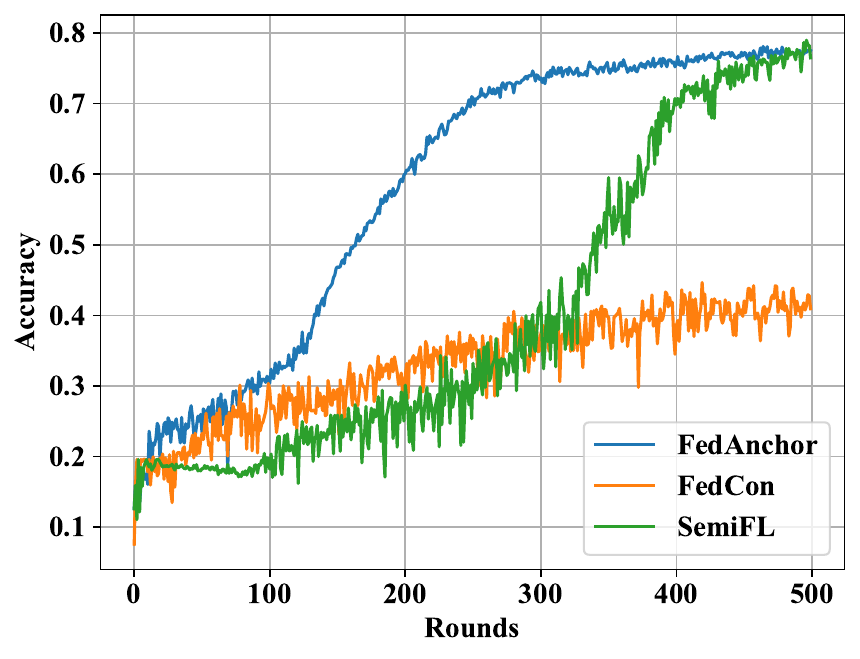}} 
    \subfigure[non-IID ($Dir=0.1$) - 1000 anchors]{\centering\includegraphics[width=0.4\textwidth]{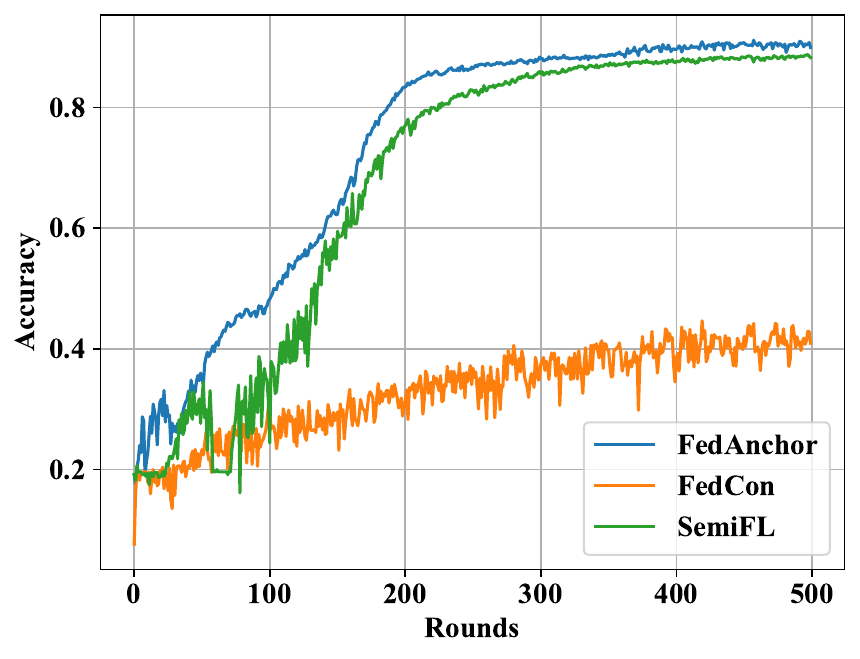}} 
\caption{\small Performance on SVHN test set on ResNet-18 with both IID and non-IID cases across different numbers of anchor data setups, compared with the baseline.}
\label{fig:acc_rounds_svhn}
\end{figure}

In this section, we provide some additional plots testing on ResNet-18 as shown in Figure \ref{fig:acc_rounds} (CIFAR10), \ref{fig:acc_rounds_cifar100} (CIFAR100) and \ref{fig:acc_rounds_svhn} (SVHN). All plots indicate that the \method outperforms the baseline significantly. FedCon consistently underperforms both \method and SemiFL. SemiFL usually has very slow convergence at the beginning of the training, as we expect, as the prediction of the model is bad at the beginning, which leads to a very limited number of training samples satisfying the threshold. We can also see that SemiFL normally generates a very unstable training curve, especially for the SVHN 250 anchors case, which also indicates the less effectiveness of using model prediction for pseudo-labeling.

\section{Pseudo-labels quality plots}\label{app:plabel_quality}

\begin{figure}[h]
    \centering
    \subfigure[CIFAR-10 non-IID ($Dir=0.1$) - 250 anchors]{\centering\includegraphics[width=0.4\textwidth]{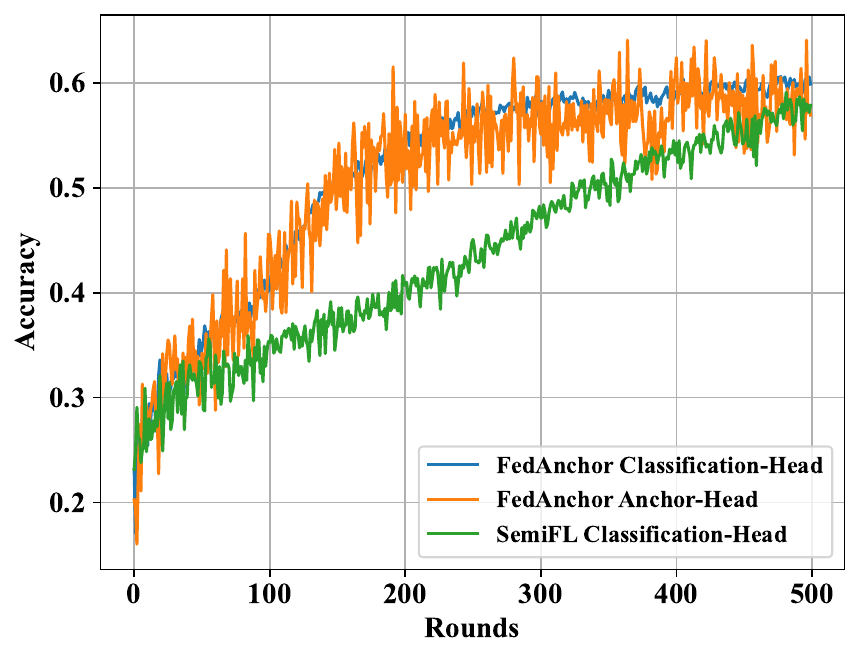}} 
    \subfigure[CIFAR-10 non-IID ($Dir=0.1$) - 500 anchors]{\centering\includegraphics[width=0.4\textwidth]{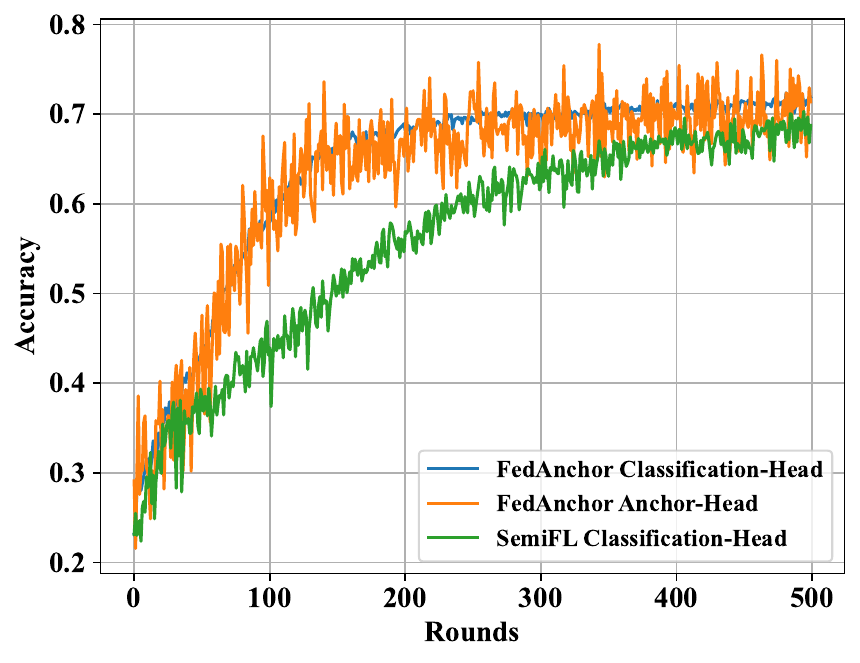}} 
    \subfigure[SVHN non-IID ($Dir=0.1$) - 250 anchors]{\centering\includegraphics[width=0.4\textwidth]{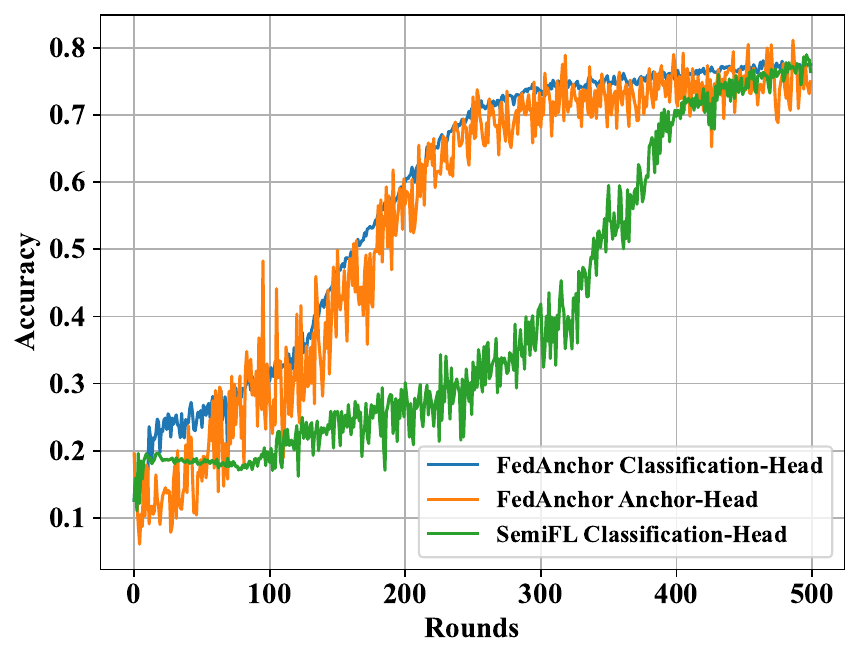}} 
    \subfigure[SVHN non-IID ($Dir=0.1$) - 100 anchors]{\centering\includegraphics[width=0.4\textwidth]{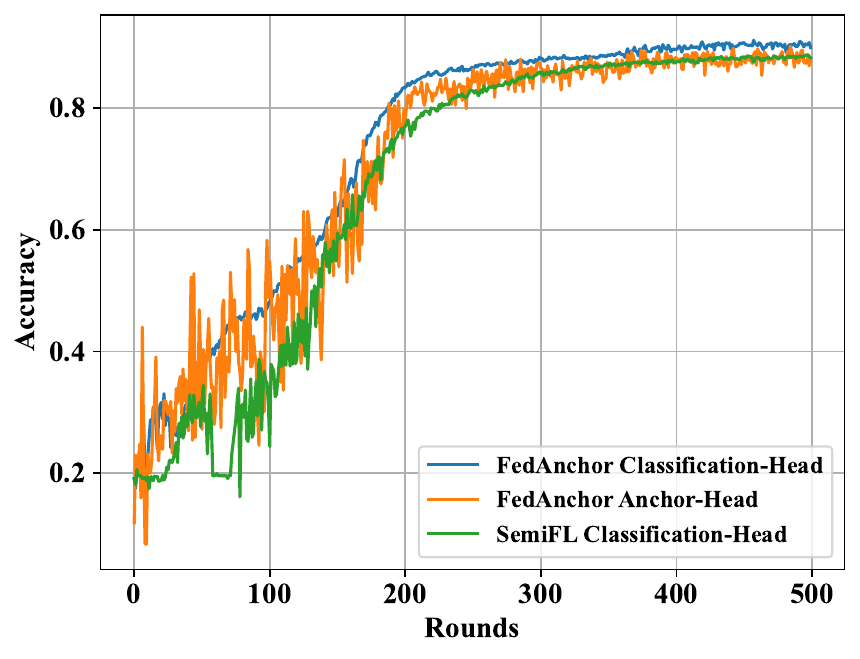}} 
\caption{\small Comparison of the pseudo-label accuracy with one of the baseline (SemiFL) pseudo-label accuracy for both CIFAR10 and SVHN with different anchor sizes. Each plot shows the pseudo-label accuracy provided by the classification head and the pseudo-label accuracy provided by our anchor head. Since SemiFL does not have the anchor head, it only has one curve for each scenario. }
\label{fig:pseudolabelquality}
\end{figure}

We compare our pseudo-label accuracy with one of the baseline (SemiFL) pseudo-label accuracy in Fig.~\ref{fig:pseudolabelquality}. We generate the plot for different datasets and anchor sizes. In each plot, we provide the pseudo-label accuracy provided by the classification head and the pseudo-label accuracy provided by our anchor head. Since SemiFL does not have the anchor head, it only has one curve for each scenario. It is clear that \method produces significantly higher pseudo-label accuracy than the baseline. Hence, \method can achieve higher performance with a faster convergence rate. Fig.~\ref{fig:pip} (right) shows the pseudo-label accuracy at the round number 100, demonstrating that the \method can produce higher pseudo-label accuracy with a big margin.

\section{Implementation and reproducibility}\label{app:impl}

During the extensive evaluation of \method, we adopted any means of making as reproducible as possible the experimental setting. Our code is publicly available in the anonymized repository at \url{https://anonymous.4open.science/r/fedanchor-8727/README.md}. The code could be easily executed in any machine possessing Nvidia GPUs. We used Poetry (\url{https://python-poetry.org/}) to create the Python package with its dependencies, such as PyTorch 2.1, Python 3.10, and Flower 1.5.0. This tool allows the researchers to reproduce the same environment that we used with minimal effort. Regarding the FedCon implementation, we followed the original repository (\url{((zewei-long/fedcon-pytorch))}).

%\section{FedCon Plot}\label{app:fedcon}
%\begin{figure}[h]
%\centering
%    \includegraphics[width=0.3\linewidth]{figures/fedcon.png}
%\caption{\small Performance on CIFAR10 test set with both IID and non-IID anchor size 5000 case for FedCon.}
%\label{fig:fedcon}
%\end{figure}

%%%%%%%%%%%%%%%%%%%%%%%%%%%%%%%%%%%%%%%%%%%%%%%%%%%%%%%%%%%%%%%%%%%%%%%%%%%%%%%
%%%%%%%%%%%%%%%%%%%%%%%%%%%%%%%%%%%%%%%%%%%%%%%%%%%%%%%%%%%%%%%%%%%%%%%%%%%%%%%

\end{document}